\documentclass[10pt,letterpaper]{article}
\usepackage[T1]{fontenc}
\usepackage{times}
\usepackage[textwidth=6.25in,textheight=9in,centering]{geometry}
\usepackage[authoryear,round]{natbib}
\usepackage{titlesec}
\titleformat{\section}{\large\bfseries}{\thesection}{0.7em}{}
\titleformat{\subsection}{\normalsize\bfseries}{\thesubsection}{0.7em}{}
\titleformat{\subsubsection}{\normalsize\bfseries}{\thesubsubsection}{0.7em}{}
\titlespacing*{\section}{0pt}{2.1ex plus .5ex minus .2ex}{1ex}
\titlespacing*{\subsection}{0pt}{1.8ex plus .4ex minus .2ex}{.8ex}
\titlespacing*{\subsubsection}{0pt}{1.6ex plus .3ex minus .2ex}{.7ex}
\IfFileExists{math_commands.tex}{

\usepackage{amsmath,amsfonts,bm}

\def\eqref#1{equation~\ref{#1}}

\def\1{\bm{1}}

\DeclareMathAlphabet{\mathsfit}{\encodingdefault}{\sfdefault}{m}{sl}
\SetMathAlphabet{\mathsfit}{bold}{\encodingdefault}{\sfdefault}{bx}{n}

}{}

\usepackage{amsmath,amssymb}
\usepackage{booktabs}
\usepackage{multirow}
\usepackage{array}
\usepackage{colortbl}
\usepackage{graphicx}
\usepackage{fontawesome5}
\usepackage{float}
\usepackage{placeins}
\usepackage{xcolor}
\usepackage{hyperref}
\hypersetup{hidelinks}
\usepackage{url}
\usepackage{xurl}
\usepackage{enumitem}
\usepackage{microtype}
\usepackage{wrapfig}
\usepackage{capt-of}
\usepackage{needspace}
\usepackage{tcolorbox}

\newtcolorbox{promptbox}{
  colback=black!4, colframe=black!70, boxrule=0.8pt,
  arc=4pt, left=10pt, right=10pt, top=7pt, bottom=7pt,
  before skip=2pt, after skip=2pt,
  fontupper=\fontencoding{T1}\fontfamily{cmtt}\fontsize{9}{11.5}\selectfont
}
\newcommand{\promptentry}[2]{%
  {\fontfamily{phv}\selectfont\bfseries #1}\par\nobreak\vspace{2pt}%
  {\raggedright #2\par}%
}
\newcommand{\promptdivider}{%
  \par\vspace{3pt}{\color{black!15}\hrule height 0.35pt}\vspace{3pt}%
}
\usepackage{xcolor}
\newcommand{\rev}[1]{#1}
\definecolor{taskChange}{RGB}{222,235,247}
\definecolor{taskSemantic}{RGB}{226,239,218}
\definecolor{taskDetection}{RGB}{252,228,214}
\definecolor{taskBuilding}{RGB}{232,225,245}
\definecolor{taskHeight}{RGB}{255,242,204}
\definecolor{taskReferring}{RGB}{218,238,243}

\title{UniFlow-RS: Unified Generative Prediction for Heterogeneous Remote-Sensing Tasks}

\author{%
Jianlong He\textsuperscript{1} \quad Liang Liao\textsuperscript{2} \quad Jing Xiao\textsuperscript{1} \quad Chaofeng Chen\textsuperscript{1,3,*} \quad Gui-Song Xia\textsuperscript{1}\\[5pt]
\normalfont\textsuperscript{1}School of Artificial Intelligence, Wuhan University\\[2pt]
\normalfont\textsuperscript{2}Hangzhou Institute of Technology, Xidian University\\[2pt]
\normalfont\textsuperscript{3}Institute for Math and AI, Wuhan
}
\date{}
\makeatletter
\renewcommand{\maketitle}{%
  \noindent\raisebox{17pt}[0pt][0pt]{\vbox{%
    \hbox to\textwidth{\includegraphics[height=23pt]{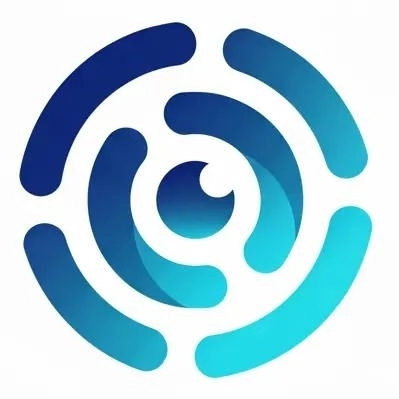}\hspace{4pt}%
      \raisebox{7pt}{\fontfamily{phv}\fontsize{10.5}{12}\selectfont\bfseries MindVision Lab}\hspace{14pt}%
      \includegraphics[trim=85bp 0bp 0bp 0bp,clip,height=23pt]{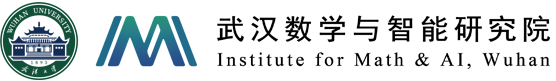}\hfill
      \includegraphics[height=23pt]{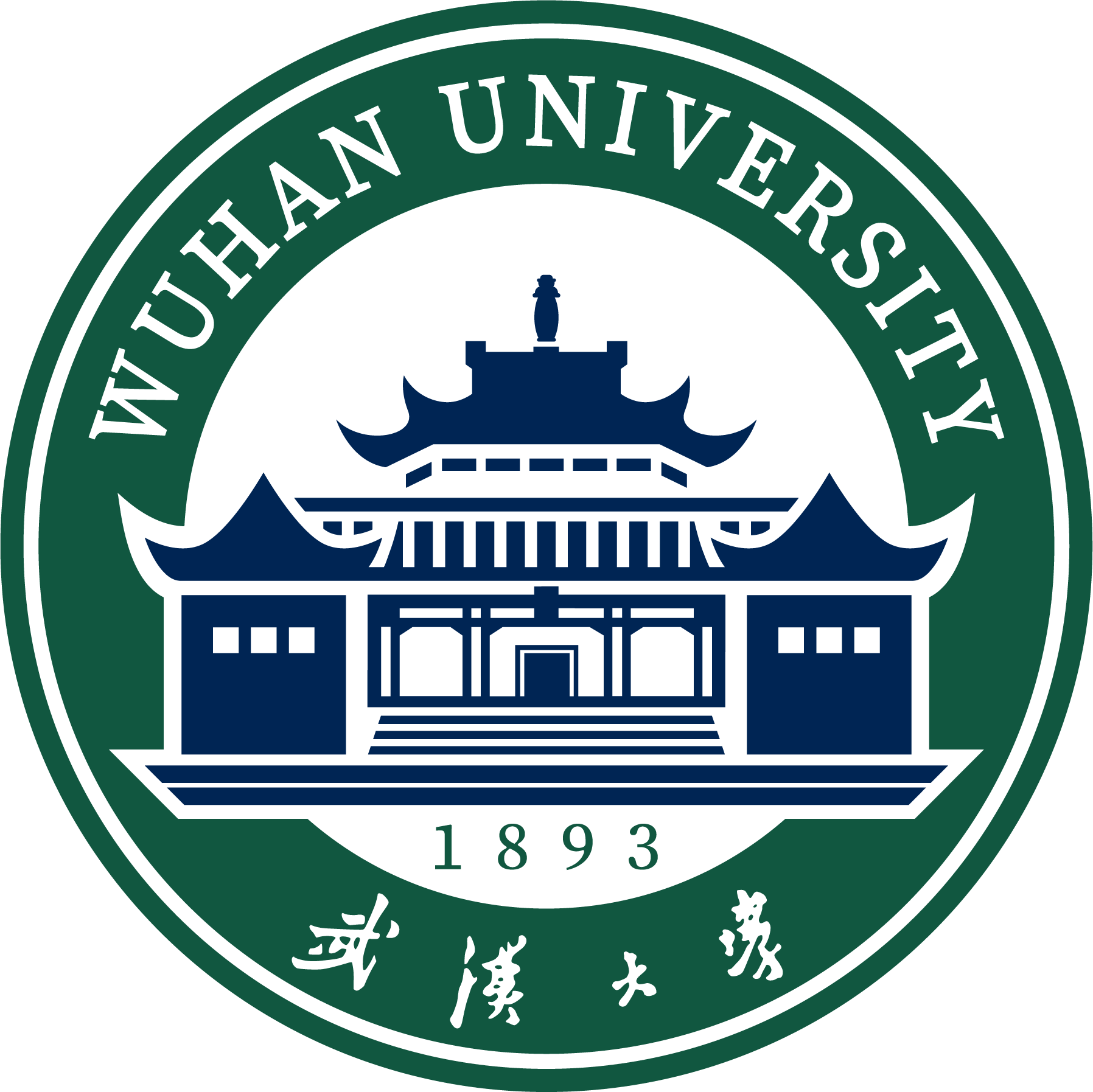}\hspace{8pt}%
      \includegraphics[height=23pt]{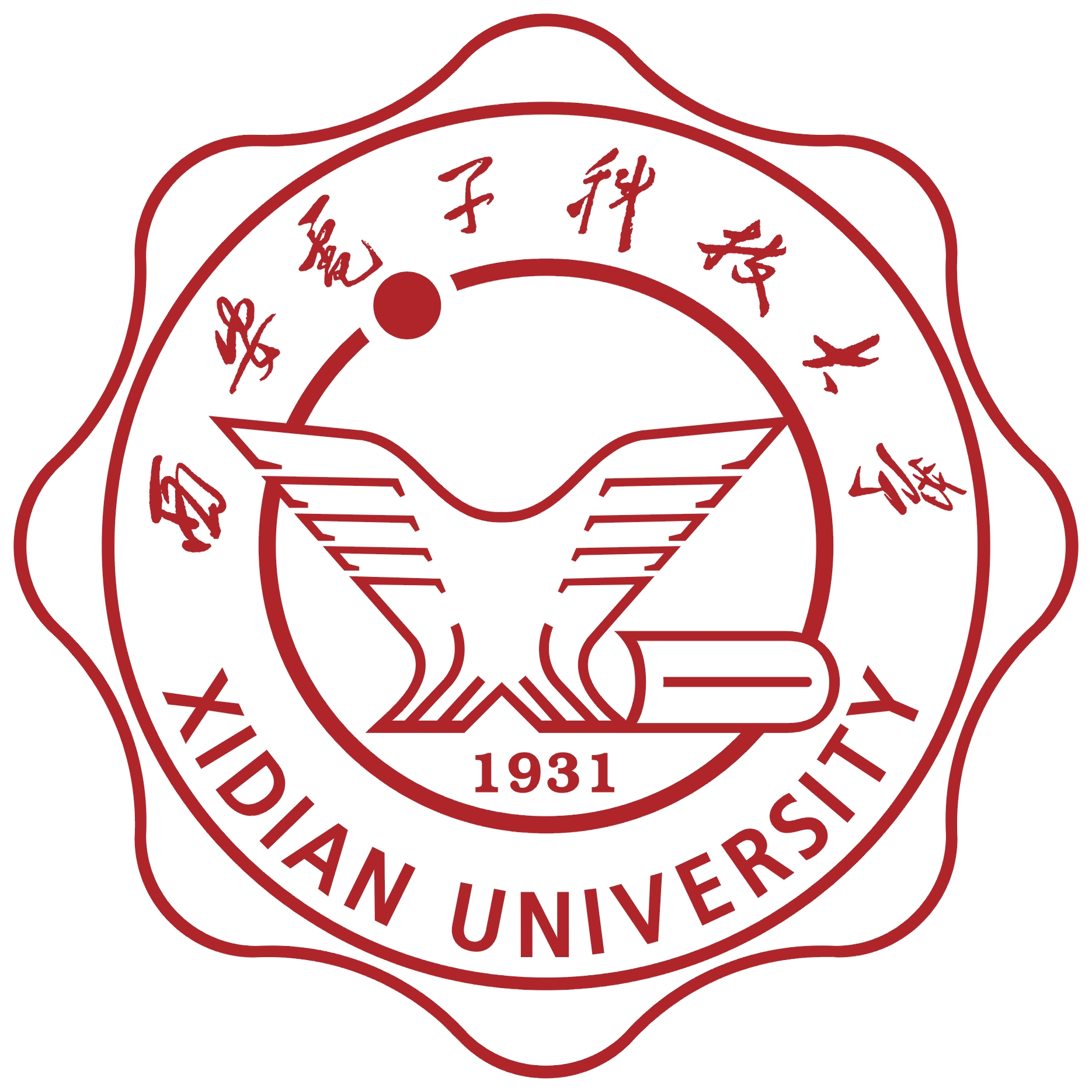}}%
    \vskip3pt{\color{black!40}\hrule height 0.4pt}%
  }}\par\vspace{-\baselineskip}
  \begin{center}
    {\fontsize{19}{22}\selectfont\bfseries\@title\par}
    \vspace{10pt}
    {\fontsize{10}{13}\selectfont\@author\par}
    \vspace{7pt}
    {\small
      \faGithub\enspace\textbf{GitHub:} \href{https://github.com/MindVision-AI/UniFlow-RS}{\textcolor{blue!70!black}{\nolinkurl{https://github.com/MindVision-AI/UniFlow-RS}}}\\[2pt]
      \faGlobe\enspace\textbf{Project Page:} \href{https://elaxegan.github.io/uniflow-rs/}{\textcolor{blue!70!black}{\nolinkurl{https://elaxegan.github.io/uniflow-rs/}}}\par}
  \end{center}
  \vspace{2pt}
}
\makeatother
\hypersetup{
  pdftitle={UniFlow-RS: Unified Generative Prediction for Heterogeneous Remote-Sensing Tasks},
  pdfauthor={Jianlong He; Liang Liao; Jing Xiao; Chaofeng Chen; Gui-Song Xia},
  pdfsubject={Research preprint}
}

\newcommand{\method}{UniFlow-RS}
\newcommand{\tmas}{TMAS}

\newcommand{\uts}{\mathrm{UTS}}
\newcommand{\Dset}{\mathcal{D}}
\newcommand{\Tset}{\mathcal{T}}
\makeatletter
\newcommand{\methodsrc}[3]{\nocite{#1}#2~{{\textcolor{gray!65}{\hyper@natlinkstart{#1}(#3)\hyper@natlinkend}}}}
\makeatother
\newcommand{\tablebodyfont}{\fontsize{7.6pt}{9pt}\selectfont}
\DeclareMathSizes{7.6}{7.6}{5}{5}

\begin{document}
\maketitle
\begingroup
\renewcommand{\thefootnote}{\fnsymbol{footnote}}
\footnotetext[1]{Corresponding author.}
\endgroup

\begin{figure}[H]
    \centering
    \includegraphics[width=\linewidth]{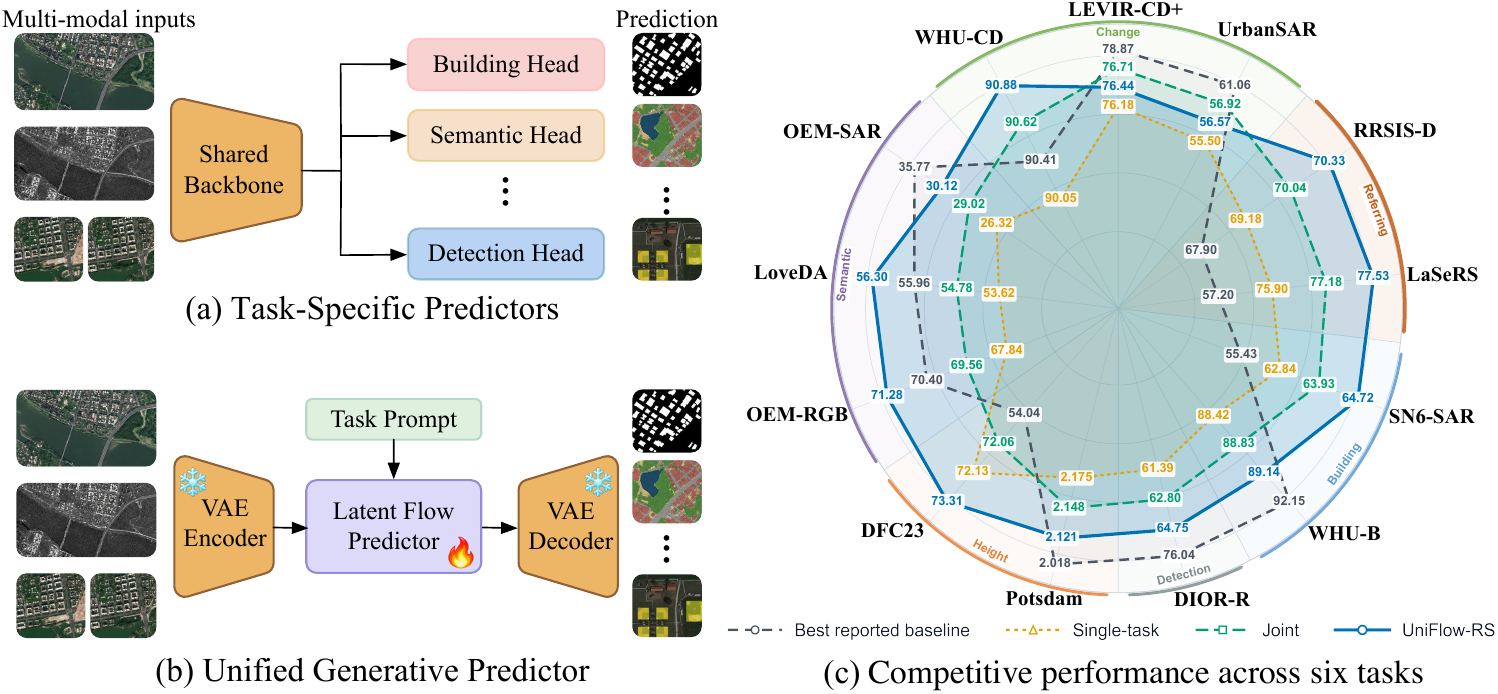}
    \caption{
    \method{} unifies heterogeneous remote-sensing perception through a shared generative prediction process. Unlike prior multi-task models that share representations but retain task-specific predictors, UniFlow-RS uses a single generative predictor across six tasks and achieves competitive performance on thirteen datasets.
    }
    \label{fig:performance}
\end{figure}

\begin{abstract}
Remote-sensing foundation models increasingly unify representations across sensing modalities and downstream tasks, yet their prediction mechanisms remain largely task-specific. Separate decoders, prediction heads, and objectives leave prediction fragmented across tasks. This naturally raises a question: \textit{can heterogeneous remote-sensing tasks share the prediction process?}
\textbf{\method} answers this question with a single latent flow-matching predictor shared across tasks. This requires addressing two key challenges: unifying heterogeneous prediction spaces and harmonizing joint optimization across tasks. First, we cast heterogeneous task outputs into structured visual targets in a common latent space, allowing them to be modeled by the same pretrained flow predictor. Task-dependent conditioning further accommodates differences in input configurations. Second, we optimize all tasks with a common Flow Matching objective, eliminating task-specific prediction losses. To mitigate optimization mismatch across tasks and modalities, we introduce Task--Modality Adaptive Timestep Sampling (TMAS), which dynamically allocates training timesteps according to the online learning state of each task--modality group.
Using a single shared model, UniFlow-RS handles six tasks and thirteen datasets, surpassing recent task-specific state-of-the-art methods on several benchmarks while remaining competitive on the others. \rev{Beyond these benchmarks, the same checkpoint demonstrates strong zero-shot transfer across unseen datasets, task settings, occlusion conditions, and near-infrared false-color imagery, without target-specific adaptation.}
Codes are available at \href{https://github.com/MindVision-AI/UniFlow-RS}{\textcolor{blue!70!black}{\nolinkurl{https://github.com/MindVision-AI/UniFlow-RS}}}.

\end{abstract}

\section{Introduction}

Remote-sensing interpretation increasingly requires models to operate across diverse tasks, sensing modalities, and input configurations. Discriminative foundation models such as SkySense and MTP ~\citep{guo2024skysense,wang2024mtp} have substantially advanced representation reuse across optical, SAR, multi-temporal observations, and downstream tasks. However, as illustrated in Figure~\ref{fig:performance}(a), most unified models still retain task-specific prediction heads, decoders, or output mechanisms after the shared representation, leaving the final prediction stage fragmented across tasks. 
This exposes a remaining gap in unified remote-sensing perception: \emph{can heterogeneous remote-sensing tasks share not only representations, but the prediction process itself?}

Generative foundation models provide a natural route toward this goal by reformulating prediction as conditional visual generation. 
Recent work has shown that pretrained image generators can serve as generalist visual predictors~\citep{gabeur2026visionbanana}, while flow matching provides a simple formulation for conditional generation through velocity-based transport~\citep{lipman2023flowmatching,liu2023rectifiedflow}. In remote sensing, generative models have further demonstrated strong flexibility across sensing modalities and generation settings~\citep{jakubik2025terramind,chen2026any2any}. 
This suggests a unified paradigm for generalist remote-sensing perception: encoding diverse structured outputs as visual representations and predicting them with a shared latent flow model.

Making this paradigm work in remote sensing requires resolving two key mismatches. The first is a \emph{prediction-space and conditioning mismatch}. Remote-sensing tasks involve heterogeneous outputs, from binary masks and semantic maps to height fields, referred regions, and rotated instances, with inputs spanning optical/SAR, single-/bi-temporal, and optional text conditioning. A unified predictor must therefore accommodate these heterogeneous output structures and input conditioning configurations without resorting to task-specific branches. The second is an \emph{optimization mismatch}. Even under a unified prediction formulation, different task--modality groups can exhibit different learning dynamics along the flow trajectory. A fixed timestep distribution imposes the same sampling schedule on groups whose optimization needs may differ substantially.

We introduce \textbf{\method}, a latent-flow framework that addresses these two mismatches with a single generative predictor shared across heterogeneous remote-sensing tasks. 
As shown in Figure~\ref{fig:performance}(b), UniFlow-RS replaces a collection of task-specific predictors with one pretrained latent flow predictor and one shared LoRA~\citep{hu2022lora} adapter, together with a frozen VAE for target encoding and decoding. 
All tasks are trained with the same Flow Matching objective, without learned task-specific output heads or task-specific losses.
To make this shared predictor applicable across tasks, UniFlow-RS maps task-specific structured targets into a common VAE-compatible visual latent space, while different tasks are specified through task-dependent input conditioning. We further design deterministic recovery rules to convert the generated latent outputs back into their native evaluation formats. 
Building on this unified prediction space, we address the heterogeneous optimization dynamics across different tasks with Task--Modality Adaptive Timestep Sampling (TMAS).
TMAS dynamically reallocates timestep-level training budget according to the online learning state of each task--modality group, allowing the sampling schedule to adapt to group-specific optimization needs.

As shown in Figure~\ref{fig:performance}(c), UniFlow-RS covers six remote-sensing tasks across thirteen datasets, spanning optical, SAR, bi-temporal, continuous-output, language-conditioned, and instance-level prediction settings. Despite using a shared LoRA adapter, UniFlow-RS surpasses the best reported results on multiple benchmarks while remaining competitive on the others. 

Our contributions are threefold:
\begin{itemize}[leftmargin=1.4em]

    \item We introduce a unified remote-sensing framework in which heterogeneous tasks share the \emph{prediction process itself}, rather than only representations. A single pretrained latent flow predictor and shared LoRA support six tasks under the same Flow Matching objective, without learned task-specific output heads or losses.
   
    \item We formulate heterogeneous remote-sensing outputs in a common structured visual space and combine it with task-dependent conditioning, allowing diverse output structures, sensing modalities, and input configurations to be handled by one latent predictor.

    \item We formulate heterogeneous task--modality learning dynamics along the flow trajectory as a timestep-allocation problem and propose TMAS, which adaptively adjusts timestep sampling based on the online learning state of each group, improving joint optimization and convergence.
\end{itemize}

\section{Related Work}

\paragraph{Unified remote-sensing models.}
Remote-sensing foundation models have increasingly moved
toward shared representations across sensing modalities
and downstream tasks.
SkySense and MTP~\citep{guo2024skysense,wang2024mtp}
establish transferable representations across heterogeneous
observations, while SkySense V2~\citep{zhang2025skysensev2}
further advances multimodal representation learning.
SelectiveMAE~\citep{wang2025selectivemae} explores
scalable masked pretraining, and
FedSense~\citep{tan2025fedsense} investigates distributed
foundation-model learning.
S5 and RingMoE~\citep{lv2026s5,bi2025ringmoe}
extend representation sharing across datasets and modalities.
Despite these advances, unification largely remains at
the representation level, with heterogeneous tasks relying
on specialized prediction heads, decoders, or objectives.
\rev{This makes the prediction interface task-dependent and
requires additional architectural design when introducing
new tasks.
In contrast, \method{} unifies the prediction process
itself by mapping heterogeneous outputs into a common
structured visual space and sharing a single generative
predictor and training objective.
This unified interface enables new dense prediction tasks
without additional learned heads, facilitating cross-task
knowledge sharing and generalization to unseen data distributions.}

\paragraph{Generative models for structured perception.}
Pretrained generative models offer transferable visual priors, making them particularly promising for remote sensing, where dense annotations are costly and scarce.
Painter~\citep{wang2023painter} unifies visual tasks through masked image modeling, while InstructDiffusion~\citep{geng2024instructdiffusion} and InstructCV~\citep{gan2024instructcv} explore instruction-conditioned diffusion for visual prediction.
FlowDIS~\citep{sargsyan2026flowdis} further applies Flow Matching to language-guided segmentation.
\rev{However, extending this paradigm to core remote-sensing perception tasks, such as building extraction, change detection, semantic segmentation, and height estimation, remains challenging due to heterogeneous output structures, sensing modalities, input configurations, and optimization dynamics.
We address these challenges through structured visual targets, observation-initialized latent flow prediction, and task--modality adaptive timestep sampling, enabling diverse remote-sensing tasks to share a unified generative prediction process.}

\paragraph{Adaptive optimization for heterogeneous joint training.}
Multi-task optimization methods such as GradNorm~\citep{chen2018gradnorm} and PIVRG~\citep{qin2025revisiting} mitigate task imbalance through gradient- or performance-based adjustments.
In generative modeling, Improved DDPM~\citep{nichol2021improved} employs loss-aware timestep sampling, while Adaptive Non-Uniform Timestep Sampling~\citep{kim2025adaptive} learns a sampling policy based on estimated objective reduction.
\rev{Heterogeneous remote-sensing perception presents a distinct challenge: diverse sensing modalities and output structures exhibit different learning dynamics along the flow trajectory of a shared predictor.
We therefore design \tmas{} specifically for heterogeneous remote-sensing joint training, adapting timestep allocation to the persistent difficulty and learning activity of each task--modality group.}

\section{\method}
\label{sec:method}

\begin{figure}[t]
    \centering
    \includegraphics[width=\linewidth]{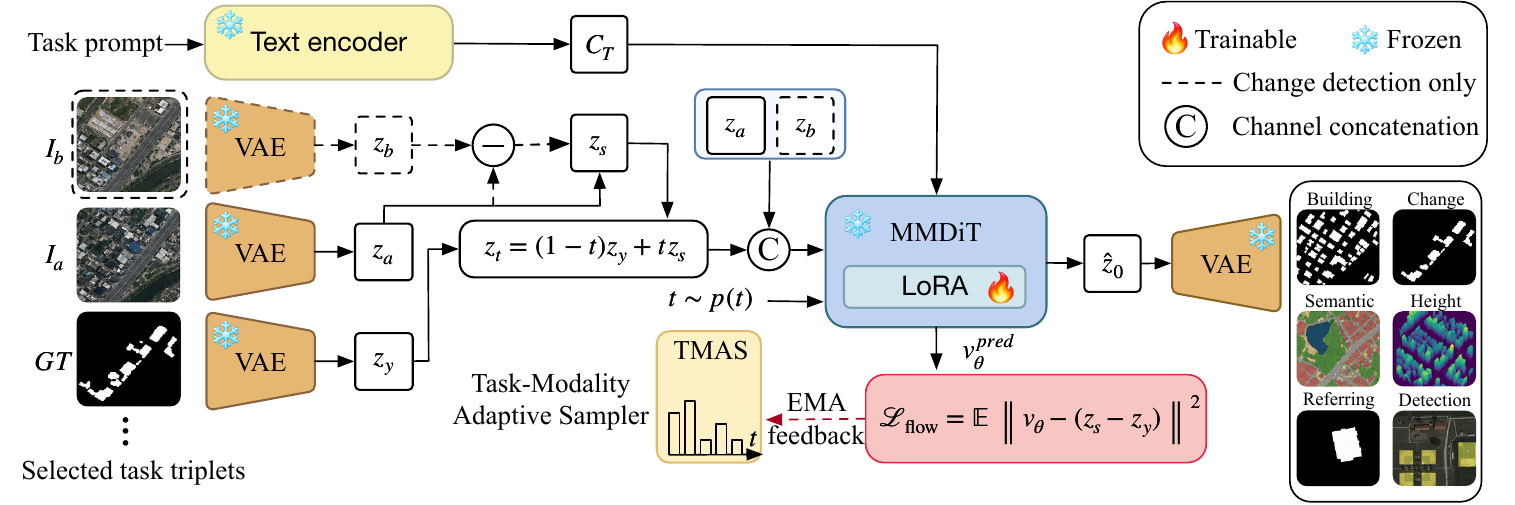}
    \caption{Overview of \method. 
    Native task outputs are rendered as structured visual targets and encoded by a frozen VAE. A pretrained MMDiT with one shared trainable LoRA predicts the conditional velocity field, while TMAS adapts timestep sampling from online Flow Matching errors. At inference, transport starts from the task-dependent source latent; the shared VAE decoder and deterministic recovery operators then produce native task predictions.
    }
    \label{fig:overview}
\end{figure}

In this section, we introduce the latent flow formulation (Sec.~\ref{sec:flow_formulation}), then specify the target representations and task-dependent conditioning (Sec.~\ref{sec:architecture}). Finally, we present \tmas{}, which adapts timestep sampling for each task--modality group (Sec.~\ref{sec:tmas}).
Figure~\ref{fig:overview} summarizes the framework.

\subsection{Unified latent flow formulation}
\label{sec:flow_formulation}

We build on a pretrained latent flow model and jointly train a single shared LoRA adapter across all tasks. The pretrained parameters of the image VAE, text encoder, and MMDiT backbone remain frozen. Let $E$, $D$, and $T$ denote the VAE encoder, VAE decoder, and text encoder, respectively, and let $\theta$ denote the trainable LoRA parameters.

For a target $y$ from task $\tau$ and dataset $d$, a deterministic rendering operator $C_{\tau,d}$ converts $y$ into a structured visual target. Its latent representation is $z_y=E(C_{\tau,d}(y))$, which allows all tasks to share the same latent prediction space without task-specific neural output heads. 
We construct the source latent $z_s$ from the input observations and connect it to the target latent $z_y$ through a linear path:
\begin{equation}
    z_t=(1-t)z_y+t z_s,
    \qquad
    v^\star=z_s-z_y,
    \qquad t\in[0,1].
    \label{eq:path}
\end{equation}
The visual condition $c_I$ and text condition $c_T$ are detailed in Sec.~\ref{sec:architecture}.

\FloatBarrier

The shared MMDiT predicts the velocity $v_\theta(z_t,c_I,c_T,t)$ with the standard flow matching objective
\begin{equation}
    \mathcal{L}_{\mathrm{flow}}
    =
    \mathbb{E}\!\left[
        \left\|v_\theta(z_t,c_I,c_T,t)-v^\star\right\|_2^2
    \right].
    \label{eq:flowloss}
\end{equation}

At inference, we solve $\frac{\mathrm{d}z_t}{\mathrm{d}t} = v_\theta\!\left(z_t,c_I,c_T,t\right)$ from $t=1$ to $t=0$ with $z_1=z_s$, obtaining $\hat z_y=z_0$.
The native prediction is recovered as $\hat y = R_d\!\left(D(\hat z_y)\right)$ where $R_d$ is the deterministic operator to convert the visual output to its native evaluation format. All tasks therefore share the latent space, velocity predictor, trainable adapter, and the frozen VAE encoder and decoder.
Appendix~\ref{app:transport_trajectories} visualizes the intermediate transport states across all six tasks.

\subsection{Structured Visual Targets and Task-Dependent Conditioning}
\label{sec:architecture}

\paragraph{Structured visual targets.}
Remote-sensing tasks have substantially different native outputs. Building extraction and referring segmentation produce binary masks; change detection includes both binary building-change masks and categorical flood-change maps; semantic segmentation produces categorical maps; height estimation predicts continuous values; and rotated detection returns oriented instances.

We encode these heterogeneous outputs as visual targets, \emph{i.e.,} images, using $C_{\tau,d}$. Binary outputs are represented as two-tone masks, categorical outputs as class-coded maps, height values through a log-compressed grayscale encoding, and rotated instances as class-coded canvases, from which the native instances are recovered via deterministic geometric post-processing. When datasets within the same task follow different label ontologies, $C_{\tau,d}$ first aligns their labels to a shared semantic definition before visual encoding.

These conversions are deterministic and non-learned, providing a common interface to the pretrained image VAE.
Detailed palettes, height transformations, ontology mappings, recovery procedures, and frozen-VAE round-trip analyses are provided in Appendix~\ref{app:target_codec}.

\paragraph{Task-dependent conditioning.}

Most tasks take a single image as input, whereas change detection uses a pair of temporal observations and referring segmentation additionally requires a language query. We accommodate these task-dependent input configurations through the source state and conditioning inputs of the same flow model. Let $I_a$ denote the current observation and $z_a=E(I_a)$ its latent representation. We use $c_T=T(q)$ to encode a fixed task instruction or, for referring segmentation, the sample's original expression (Appendix~\ref{app:text_prompts}). For bi-temporal tasks, the previous observation $I_b$ is additionally encoded as $z_b=E(I_b)$. The source state and visual condition are
\begin{equation}
[z_s,c_I]=
\begin{cases}
    \left[z_a,\,[\mathbf{0},z_a]\right],
    & \text{single-image tasks},\\[1mm]
    \left[z_a-z_b,\,[z_b,z_a]\right],
    & \text{bi-temporal tasks},
\end{cases}
\label{eq:source}
\end{equation}
where $[\cdot,\cdot]$ denotes channel-wise concatenation, and
$\mathbf{0}$ has the same shape as $z_a$.
The source state $z_s$ initializes the flow, while $c_I$ provides visual context.
We use $z_a$ as the source for single-image tasks to preserve input structure,
and $z_a-z_b$ for change detection while conditioning on $[z_b,z_a]$ to retain both observations.
We validate these choices in Sec.~\ref{sec:source_ablation}.

\subsection{Task--Modality Adaptive Timestep Sampling}
\label{sec:tmas}

The formulation above unifies heterogeneous tasks under a shared latent predictor and Flow Matching objective, but it does not eliminate differences in their optimization dynamics.

In particular, different tasks and modalities may exhibit distinct learning difficulties along the flow trajectory due to differences in output structure and sensing characteristics. 

A single global timestep distribution may oversample regions that are already well learned for some groups while undersampling more challenging regions for others.

\begin{figure}[!t]
    \centering
    \includegraphics[width=\linewidth]{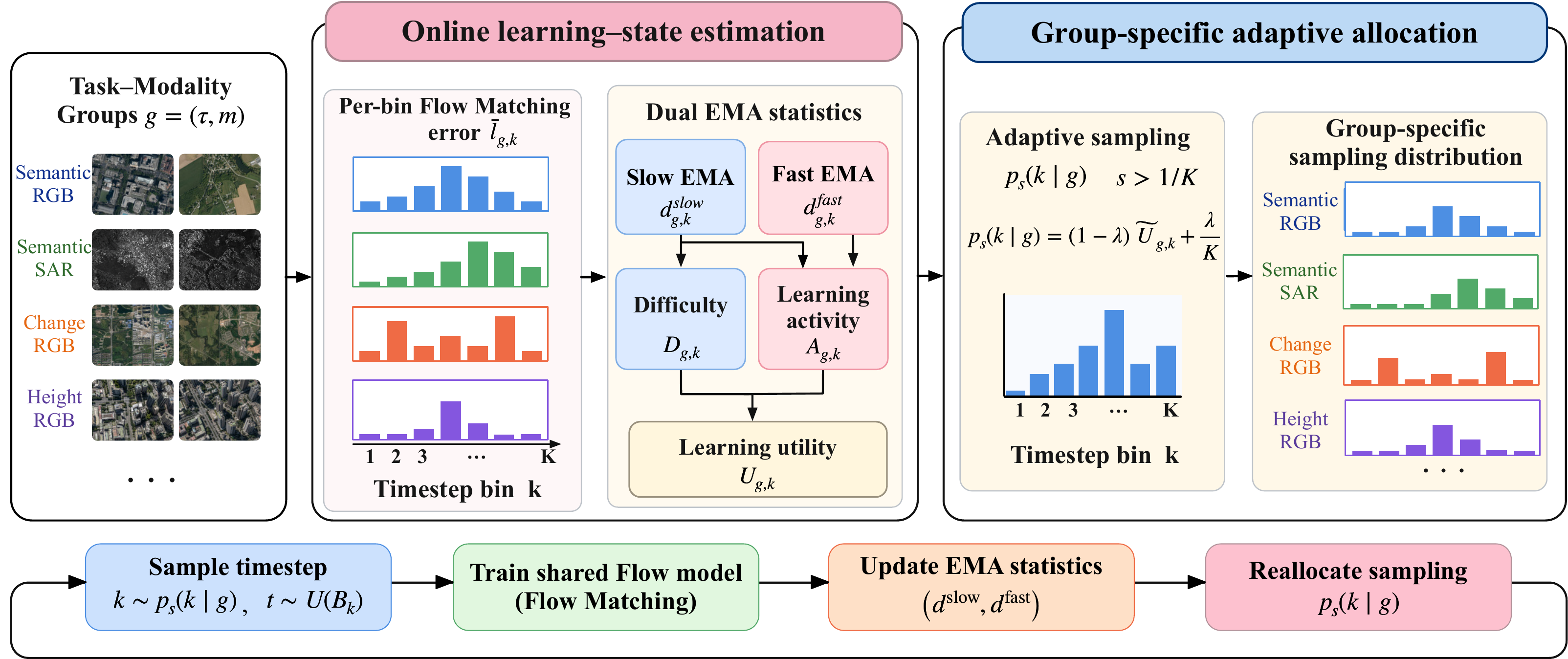}
    \caption{Task--Modality Adaptive Timestep Sampling. Each group is indexed by task and modality. TMAS tracks group-specific Flow Matching error at two timescales, converts the resulting learning state into a sampling utility, and adjusts the timestep distribution after uniform warm-up. A persistent uniform component ensures continued exploration across the flow trajectory.}
    \label{fig:tmas}
\end{figure}

A simple solution is to sample more often from regions with larger prediction errors.
However, error magnitude alone does not tell us how a region is being learned.
A high-error region may have plateaued, whereas a moderate-error region may still be changing rapidly.
We therefore use two signals: \emph{persistent difficulty}, which describes where a group remains difficult, and \emph{learning activity}, which describes where its error is still changing.
\tmas{} estimates these signals for each task--modality group and uses them to adapt timestep sampling online.
Figure~\ref{fig:tmas} gives an overview.

\FloatBarrier

\paragraph{Online learning-state estimation.}

We define each task--modality group as $g=(\tau,m)$, where $\tau$ denotes the task and $m$ the physical modality. 
Datasets with the same task and modality share one sampling profile.
We divide $[0,1]$ into $K$ equal-width bins and let $\bar{\ell}_{g,k}$ be the mean normalized velocity-prediction error observed for group $g$ at bin $k$.

To characterize both persistent difficulty and recent learning dynamics, we maintain fast and slow exponential moving averages (EMAs) of the bin-wise error:
\begin{equation}
    \begin{aligned}
    d^{\mathrm{fast}}_{g,k}
    &\leftarrow \left(1-\frac{1}{K}\right)d^{\mathrm{fast}}_{g,k}
    +\frac{1}{K}\bar\ell_{g,k},&
    d^{\mathrm{slow}}_{g,k}
    &\leftarrow \left(1-\frac{1}{K^2}\right)d^{\mathrm{slow}}_{g,k}
    +\frac{1}{K^2}\bar\ell_{g,k}.
    \end{aligned}
    \label{eq:difficulty}
\end{equation}
The slow EMA captures the long-term difficulty of a bin, while the fast EMA is more sensitive to recent updates.
Their discrepancy therefore reflects ongoing learning dynamics: $d^{\mathrm{fast}}_{g,k}<d^{\mathrm{slow}}_{g,k}$ suggests recent improvement, while $d^{\mathrm{fast}}_{g,k}>d^{\mathrm{slow}}_{g,k}$ reflects increasing error
which can indicate forgetting or cross-task interference.

Implementation details including error normalization, initialization, aggregation, and unseen-bin handling are given in Appendix~\ref{app:tmas_details}.

Absolute Flow Matching errors can have different scales across task--modality groups.
We therefore measure difficulty relative to other timestep bins of the same group.
Before computing the sampling scores, we convert the raw EMA states into stable
sampling-time estimates: unobserved bins are assigned a conservative group-level
fallback, and all estimates are lower-bounded by $\epsilon$ for numerical stability.
We denote the resulting estimates by
$\tilde d^{\mathrm{slow}}_{g,k}$ and $\tilde d^{\mathrm{fast}}_{g,k}$;
the exact fallback rule is given in Appendix~\ref{app:tmas_details}.
We then define
\begin{equation}
\begin{aligned}
D_{g,k}
&=
\frac{\tilde d^{\mathrm{slow}}_{g,k}}
{\frac{1}{K}\sum_{j=1}^{K}\tilde d^{\mathrm{slow}}_{g,j}},&
A_{g,k}
&=
\min\!\left\{
\left|
\frac{\tilde d^{\mathrm{slow}}_{g,k}-\tilde d^{\mathrm{fast}}_{g,k}}
{\tilde d^{\mathrm{slow}}_{g,k}}
\right|,
1
\right\}.
\end{aligned}
\label{eq:learning_state}
\end{equation}

Here, $D_{g,k}>1$ indicates that bin $k$ is more difficult than the group average, while $A_{g,k}$ measures how strongly its error is changing relative to its long-term level. We use the absolute fast--slow discrepancy because both rapid improvement and rapid degradation indicate active learning dynamics.

We combine them into
\begin{equation}
U_{g,k} =
(D_{g,k}+\epsilon)
\left(\frac{1}{K}+A_{g,k}\right),
\label{eq:utility}
\end{equation}
where $\epsilon>0$ ensures numerical stability.
Difficulty provides the base allocation, while activity gives additional weight to bins whose learning state is changing.
The offset $1/K$ ensures that a difficult but temporarily stable bin still receives training budget when the fast and slow estimates are close.

\paragraph{Group-specific adaptive allocation.}

The estimates are unreliable early in training because many group--bin pairs have only been observed a few times.
We therefore use uniform timestep sampling for the first $1/K$ fraction of training.
Let $s\in[0,1]$ denote normalized training progress.
The probability of sampling bin $k$ for group $g$ is
\begin{equation}
p_s(k\mid g) =
\begin{cases}
    \dfrac{1}{K},
    & s\leq \dfrac{1}{K}, \\[2mm]
    \dfrac{\lambda}{K}
    +(1-\lambda)
    \dfrac{U_{g,k}}{\sum_{j=1}^{K}U_{g,j}},
    & s> \dfrac{1}{K},
\end{cases}
\label{eq:tmas}
\end{equation}
where $\lambda\in(0,1]$ controls the trade-off between uniform exploration and adaptive sampling.
After selecting a bin, $t$ is sampled uniformly within it. The uniform term $\lambda/K$ prevents a bin that is underestimated early from being permanently neglected.

TMAS thus assigns different timestep distributions to different
task--modality groups while keeping the task and dataset
sampling proportions unchanged.

\section{Experiments}
\label{sec:experiments}

\subsection{Experimental setup}
\label{sec:implementation}
We use the frozen FLUX.2 [klein] 4B Base MMDiT~\citep{bfl2026flux2klein}, together with its frozen VAE and a frozen Qwen3-4B-FP8 text encoder~\citep{qwen2025qwen3}. A  rank-32 LoRA with approximately 49.1M trainable parameters is shared across all tasks. We train for 60k optimizer steps with batch size 32 on four NVIDIA RTX 6000 Ada GPUs using AdamW and an initial learning rate of $5\times10^{-5}$, halved at 20k, 35k, and 50k steps.  Unless otherwise specified, inference uses two flow steps. \tmas{} uses $K=10$ timestep bins and exploration weight $\lambda=0.2$; all remaining sampler coefficients are fixed analytically from $K$, with $\epsilon=10^{-6}$ for numerical stability.

We evaluate six tasks across thirteen datasets,
following the official splits and native metrics
whenever available. We further assess zero-shot generalization on five
additional datasets across three settings.
All evaluations use the same jointly trained
checkpoint without target-specific adaptation.
Full details of the datasets, preprocessing, and
evaluation metrics are provided in the appendix.
For controlled comparisons among unified model
variants, we additionally report unified task score
(UTS), which first averages the eligible native
metrics within each task and then assigns equal
weight to all six training tasks.
Zero-shot transfer results are reported separately
and are not included in UTS.

\nocite{zhang2026unichange,guo2024skysense,wang2024mtp,wang2025selectivemae,
tan2025fedsense,lv2026s5,wang2024metasegnet,
zhou2026tricdnet,dong2024changeclip,cheng2022mask2former,zou2025d2ls,
xia2023openearthmap,xia2025openearthmapsar,bi2025ringmoe,
zhao2025ccs,he2024uanet,wang2024bidnet,pei2026aemkd,chen2024heightformer,
lei2025raephrnet,mao2023light,yang2026mars,liu2024rmsin,
rong2026rs2sam2,liu2026visionreasoner,shabbir2025geopixel,xin2026segearthr2,
li2025uabcd,li2025segearthov,huang2024mutdet,lee2025hausdorff,
jang2025mmr,du2024lumnet,wei2026cmhcl,yao2026uaglnet,
shi2021dmbc,xiong2024dofa,yang2026changerwkv}

\begin{figure}[!t]
    \centering
    \includegraphics[width=\linewidth]{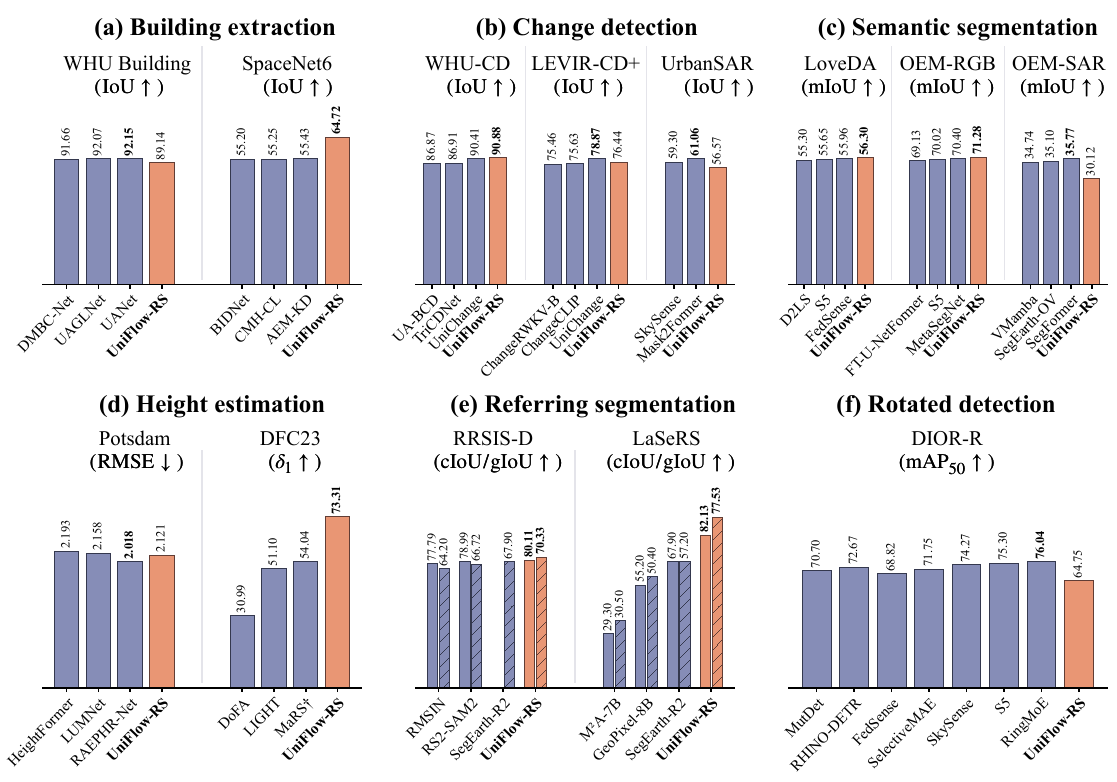}
    \caption{
Main benchmark comparison across six tasks and thirteen datasets.
Each panel reports the dataset-native metric indicated above the bars, with arrows showing the preferred direction. All \method{} results are obtained from the same shared checkpoint.
}
    \label{fig:main_results}
\end{figure}

\subsection{Main results}
\label{sec:main_results}

Figure~\ref{fig:main_results} compares \method{} with representative unified and task-specific baselines across all thirteen datasets. All \method{} results are obtained from a single jointly trained checkpoint with a shared LoRA, without task-specific fine-tuning or learned output heads. Full numerical comparisons, venue--year information, and dataset-specific notes are provided in Appendix~\ref{app:full_benchmarks} (Tables~\ref{tab:main_a} and~\ref{tab:main_b}).

\method{} achieves the best results in our comparison on WHU-CD~\citep{ji2019whu}, LoveDA~\citep{wang2021loveda}, OpenEarthMap-RGB~\citep{xia2023openearthmap}, SpaceNet~6 SAR~\citep{shermeyer2020spacenet6}, DFC23~\citep{liu2024dfc23}, RRSIS-D~\citep{liu2024rmsin}, and LaSeRS~\citep{xin2026segearthr2}, while maintaining competitive performance across all six tasks. These results demonstrate the effectiveness of a shared generative prediction process across diverse sensing modalities and output structures.

The task-wise results also reflect different compatibility with the unified
prediction interface. Referring segmentation benefits from the shared dense
prediction prior and explicit language conditioning, making its binary-mask
output well aligned with conditional latent transport. In contrast, rotated
detection requires recovering discrete instances and precise oriented geometry
from a generated class-coded canvas. Without task-specific box or orientation
losses, small generation errors can be amplified during deterministic instance
recovery, making detection more challenging.

\begin{figure}[!t]
\centering
\includegraphics[width=\linewidth]{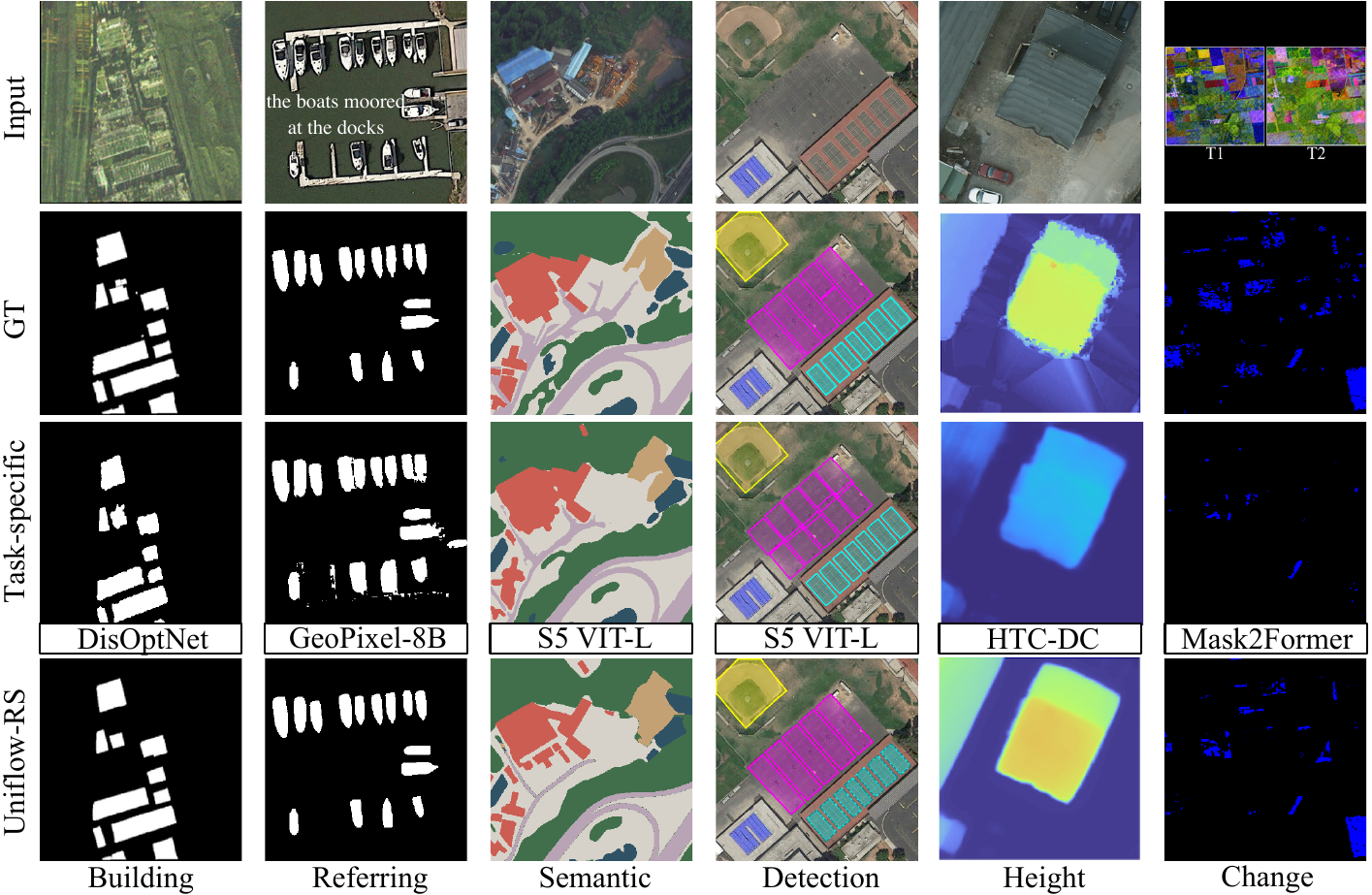}
\caption{
Qualitative comparison of task-specific baselines and \method{} across six tasks.
Baseline names are shown below the corresponding task-specific predictions.
}
\label{fig:qualitative_main}
\end{figure}


\paragraph{Qualitative results.}
Figure~\ref{fig:qualitative_main} compares the shared
\method{} predictor with task-specific baselines
~\citep{disoptnet,shabbir2025geopixel,lv2026s5,htcdc,cheng2022mask2former}
across six tasks.
Overall, \method{} produces cleaner predictions with
fewer spurious regions, more coherent semantic boundaries,
and better-preserved spatial structures.
In change detection, it recovers more change regions
while suppressing isolated responses, whereas in rotated
detection, it produces more coherent instance layouts.

Notably, in building extraction and height estimation,
\method{} generates smoother and more spatially continuous
predictions that better follow the underlying scene structures.
Although some predictions do not perfectly match the
ground-truth annotations, they often remain visually
plausible and consistent with human perception of the scene.
These observations highlight the potential of generative
prediction to capture meaningful spatial priors and
produce structurally coherent outputs, rather than
merely reproducing annotation patterns.
Additional qualitative comparisons across all six tasks
are provided in Appendix~\ref{app:additional_qualitative}.

\subsection{Analysis}
\label{sec:analysis}

\begin{table}[!htbp]
\caption{Comparison of task-specific and joint adaptation. Single-task uses one LoRA per task, whereas the joint variants share one LoRA.}
\label{tab:joint}
\centering
\tablebodyfont
\setlength{\tabcolsep}{4.0pt}
\renewcommand{\arraystretch}{1.08}
\begin{tabular}{lccccccc}
\toprule
Setting
& \shortstack{Building\\IoU $\uparrow$}
& \shortstack{Change\\IoU $\uparrow$}
& \shortstack{Semantic\\mIoU $\uparrow$}
& \shortstack{Height\\$\delta_1 \uparrow$}
& \shortstack{Ref.\\gIoU $\uparrow$}
& \shortstack{Det.\\mAP50 $\uparrow$}
& UTS $\uparrow$ \\
\midrule
Single-task
& 75.63 & 73.91 & 49.26 & 72.13 & 72.54 & 61.39 & 67.48 \\
Joint
& 76.38 & \textbf{74.75} & 51.12 & 72.06 & 73.61 & 62.80 & 68.45 \\
\textbf{Joint + \tmas{}}
& \textbf{76.93} & 74.63 & \textbf{52.57}
& \textbf{73.31} & \textbf{73.93} & \textbf{64.75}
& \textbf{69.35} \\
\bottomrule
\end{tabular}
\end{table}

\paragraph{Joint training vs. task-specific training.}
\label{sec:generalist}
Table~\ref{tab:joint} compares a single jointly trained model with six independently trained task-specific models, all using the same Fixed Beta sampler. Joint training increases UTS from 67.48 to 68.45, with improvements on five of the six tasks. This suggests that heterogeneous tasks can benefit from the joint training process which includes more diverse datasets. With \tmas{}, the jointly trained model further improves UTS to 69.35.

\begin{table}[!htbp]
\caption{
\rev{Comparison of timestep-sampling strategies under joint training.
All results are measured after 60k updates.
Steps denotes the first update reaching the final UTS of Fixed Beta, NR means not reached.}
}
\label{tab:sampler}
\centering
\tablebodyfont
\setlength{\tabcolsep}{4.0pt}
\begin{tabular}{lcccccccc}
\toprule
Sampler
& \shortstack{Building\\IoU $\uparrow$}
& \shortstack{Change\\IoU $\uparrow$}
& \shortstack{Semantic\\mIoU $\uparrow$}
& \shortstack{Height\\$\delta_1 \uparrow$}
& \shortstack{Ref.\\gIoU $\uparrow$}
& \shortstack{Det.\\mAP50 $\uparrow$}
& UTS $\uparrow$
& \shortstack{Steps $\downarrow$} \\
\midrule

Uniform
& 75.92 & 74.42 & 50.31 & 71.84
& 73.20 & 61.95 & 67.94 & NR \\

Fixed Beta $(2.5,1)$
& 76.38 & \textbf{74.75} & 51.12 & 72.06
& 73.61 & 62.80 & 68.45 & 60k \\

\rev{Adaptive non-uniform}
& 75.98 & 74.36 & 50.47 & 72.13
& 73.72 & 62.20 & 68.14 & 58k \\

\midrule

Global adaptive
& 76.54 & 74.61 & 51.68 & 72.33
& 73.52 & 63.03 & 68.62 & 53k \\

\tmas{} w/o activity
& 76.74 & 74.48 & 52.23 & 72.96
& 73.87 & 63.88 & 69.03 & 44k \\

\textbf{\tmas{}}
& \textbf{76.93} & 74.63
& \textbf{52.57} & \textbf{73.31}
& \textbf{73.93} & \textbf{64.75}
& \textbf{69.35} & \textbf{37k} \\

\bottomrule
\end{tabular}
\end{table}

\paragraph{Task--modality adaptive timestep sampling.}
Table~\ref{tab:sampler} compares fixed and adaptive timestep-sampling
strategies under the same joint-training protocol.
Compared with Fixed Beta (68.45 UTS), global adaptive sampling
improves UTS to 68.62, while introducing task--modality-specific
difficulty profiles increases it to 69.03.
Incorporating learning activity raises UTS to 69.35,
with the largest gains over Fixed Beta in rotated detection
(+1.95), semantic segmentation (+1.45), and height estimation
(+1.25).
\rev{\tmas{} reaches the final performance of Fixed Beta
in 37k updates, requiring 38.3\% fewer training steps.
In comparison, Adaptive Non-Uniform Timestep Sampling
~\citep{kim2025adaptive} achieves a final UTS of 68.14,
despite first reaching the Fixed Beta reference at 58k updates.}
These results demonstrate the benefits of adapting timestep
allocation to the difficulty and learning activity of
task--modality groups.
Additional component ablations are provided in
Appendix~\ref{app:tmas_components}.

\par\addvspace{6pt}
\noindent\begin{minipage}[t]{0.47\textwidth}
\vspace{0pt}
Figure~\ref{fig:dynamics} illustrates the timestep-wise
difficulty profiles across task--modality groups after
normalizing out the global error trend.
SAR building extraction and semantic segmentation
exhibit higher relative errors at larger timesteps,
whereas RGB change detection and rotated detection
show greater difficulty near the target end of the
flow trajectory.
These distinct profiles motivate group-specific
timestep allocation.
Unlike a global sampler that uses a shared
distribution across tasks, \tmas{} adapts the
sampling distribution to each group's learning
state, consistent with its improvements over
fixed and global sampling in
Table~\ref{tab:sampler}.
\end{minipage}\hfill
\begin{minipage}[t]{0.5\textwidth}
\vspace{0pt}
\centering
\includegraphics[width=\linewidth]{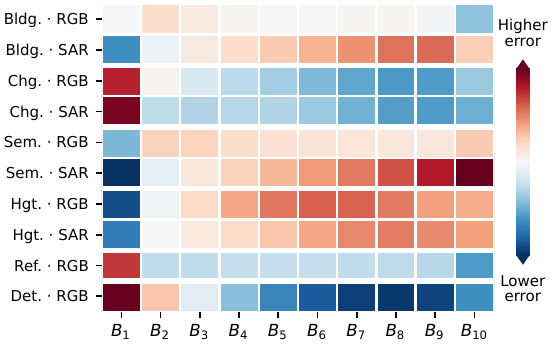}
\captionof{figure}{Task--modality timestep difficulty.}
\label{fig:dynamics}
\end{minipage}
\par\addvspace{8pt}

\begin{table}[H]
\caption{Zero-shot transfer to unseen datasets under visual and visual--language
out-of-distribution (OOD) shifts. EarthReason includes both visual and language shifts,
whereas RISBench mainly introduces a visual shift. All \method{} results use the same jointly
trained checkpoint.}
\label{tab:generalization}
\centering
\tablebodyfont
\setlength{\tabcolsep}{3.8pt}
\renewcommand{\arraystretch}{1.06}

\begin{tabular*}{\linewidth}{@{\extracolsep{\fill}}lcc@{\hspace{1.0em}}lcc}
\toprule
\multicolumn{3}{c}{\textbf{EarthReason} (Visual+Language OOD)}
&
\multicolumn{3}{c}{\textbf{RISBench} (Visual OOD)}
\\
\cmidrule(lr){1-3}\cmidrule(lr){4-6}

Method & cIoU $\uparrow$ & gIoU $\uparrow$
& Method & cIoU $\uparrow$ & gIoU $\uparrow$ \\
\midrule

FLUX.2-Klein Base
& 17.74 & 17.04
& FLUX.2-Klein Base
& 6.38 & 7.74 \\

\methodsrc{rong2026rs2sam2}{RS2-SAM2-RRSISD}{AAAI'26}
& 35.48 & 36.55
& \methodsrc{liu2026visionreasoner}{VisionReasoner}{ICLR'26}
& 46.98 & 51.62 \\

\textbf{\method{}}
& \textbf{44.23} & \textbf{48.24}
& \textbf{\method{}}
& \textbf{70.92} & \textbf{64.62} \\

\bottomrule
\end{tabular*}
\end{table}

\paragraph{Generalization beyond the training datasets.}
\label{sec:transfer}

We evaluate the same jointly trained \method{} checkpoint on two unseen referring-segmentation datasets without target-specific adaptation. RISBench mainly introduces a visual domain shift while retaining explicit referring expressions, whereas EarthReason additionally contains more implicit, reasoning-oriented language queries. As shown in Table~\ref{tab:generalization}, \method{} achieves 70.92/64.62 cIoU/gIoU on RISBench and 44.23/48.24 on EarthReason, outperforming the corresponding baselines by 23.94/13.00 and 8.75/11.69 points, respectively. The results show that the shared predictor generalizes to unseen visual and language distributions without additional adaptation.


\begin{table}[H]
\caption{
\rev{Zero-shot transfer to unseen tasks under occlusion.
Chesapeake RSC evaluates road extraction under tree-canopy occlusion
(IoU, F1, and canopy-road recall $R_{\rm canopy}$),
while PIE evaluates four-class semantic segmentation under
clear and synthetic-cloud conditions.
All models are evaluated without target-domain adaptation.}
}
\label{tab:rsc_occlusion}
\centering
\tablebodyfont
\setlength{\tabcolsep}{0pt}
\renewcommand{\arraystretch}{1.12}
\begin{tabular}{@{}>{\raggedright\arraybackslash}p{0.21\linewidth}
*{3}{>{\centering\arraybackslash}p{0.09\linewidth}}
p{0.04\linewidth}
>{\raggedright\arraybackslash}p{0.21\linewidth}
*{3}{>{\centering\arraybackslash}p{0.09\linewidth}}@{}}
\toprule
\multicolumn{4}{c}{\textbf{Chesapeake RSC} (Canopy occlusion)}
& & \multicolumn{4}{c}{\textbf{PIE} (Cloud occlusion)} \\
\cmidrule(lr){1-4}\cmidrule(lr){6-9}
Method & IoU $\uparrow$ & F1 $\uparrow$ & $R_{\rm canopy}\uparrow$
& & Method & Clear & Thin & Thick \\
\midrule
FLUX.2-Klein Base & 13.05 & 23.08 & 17.22
& & FLUX.2-Klein Base & 9.73 & 10.32 & 7.16 \\
\methodsrc{lv2026s5}{S5 ViT-L}{AAAI'26} & 37.40 & 54.43 & 13.71
& & \methodsrc{lv2026s5}{S5 ViT-L}{AAAI'26} & 36.62 & 14.17 & 4.76 \\
\methodsrc{peng2026roadgie}{RoadGIE}{CVPR'26} & 50.49 & 67.10 & 29.75
& & \methodsrc{wei2026mmovseg}{MM-OVSeg}{CVPR'26} & 24.07 & 19.82 & 16.61 \\
\textbf{\method{}} & \textbf{52.20} & \textbf{68.59} & \textbf{38.89}
& & \textbf{\method{}} & \textbf{38.92} & \textbf{32.21} & \textbf{24.52} \\
\bottomrule
\end{tabular}
\end{table}

\rev{\paragraph{Zero-shot transfer to unseen tasks.}
We further evaluate the same jointly trained \method{}
checkpoint on two unseen tasks without target-specific adaptation.
As shown in Table~\ref{tab:rsc_occlusion},
\method{} achieves 52.20 IoU and 38.89 canopy-road recall
on Chesapeake RSC, outperforming the automatic RoadGIE
baseline by 1.71 and 9.14 points, respectively.
On PIE, \method{} achieves 38.92, 32.21, and 24.52 mIoU
under clear, thin-cloud, and thick-cloud conditions.
Compared with S5, the improvements increase from
2.30 points under clear conditions to 18.04 and
19.76 points under thin and thick clouds, respectively.
These results demonstrate the ability of the shared generative
predictor to generalize to unseen task settings and maintain
robust performance under visual occlusion.
Detailed evaluation protocols are provided in
Appendix~\ref{sec:rsc_protocol}.}

\Needspace{13\baselineskip}
\begin{wrapfigure}{r}{0.50\textwidth}
    \centering
    \vspace{-3mm}
    \includegraphics[width=0.5\textwidth]
    {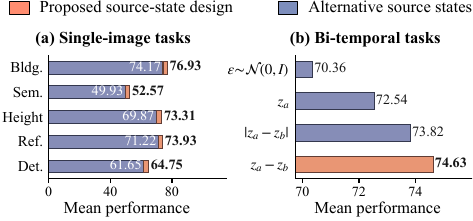}
    \vspace{-2mm}
    \caption{Source-state initialization ablation.}
    \label{fig:source_state}
    \vspace{-3mm}
\end{wrapfigure}

\paragraph{Source-state initialization.}
\label{sec:source_ablation}
We isolate the source-state design in Eq.~\ref{eq:source} under fixed visual
conditioning. Figure~\ref{fig:source_state}(a) shows that $z_a$ improves all
five single-image tasks over Gaussian initialization by 2.64--3.44 points,
confirming the benefit of preserving input structure. For bi-temporal change
detection, the mean score increases from 70.36 with Gaussian initialization
to 72.54 with $z_a$, 73.82 with $|z_a-z_b|$, and 74.63 with $z_a-z_b$,
showing that preserving directional temporal differences is beneficial.

\section{Conclusion}


We presented \method{}, a unified latent-flow framework for heterogeneous remote-sensing perception. Instead of sharing only representations while retaining task-specific predictors, \method{} unifies the prediction process through a pretrained flow model with a jointly trained LoRA. Structured visual targets and task-dependent conditioning enable six tasks across thirteen datasets to share the same latent prediction space and Flow Matching objective, without learned task-specific output heads or losses. To better optimize this shared predictor, \tmas{} adapts timestep sampling to the online learning state of each task--modality group.

\rev{Across a diverse set of benchmarks, the resulting model remains competitive with task-specific methods, benefits from joint training, and demonstrates strong zero-shot generalization across unseen datasets, task settings, and sensing conditions without target-specific adaptation}. Overall, our results demonstrate the potential of a shared generative prediction process as a foundation for generalist remote-sensing perception, and motivate further research toward more scalable and broadly transferable unified models.

\begingroup
\hypersetup{hidelinks}
\renewcommand{\bibfont}{\normalfont\fontsize{10pt}{11pt}\selectfont}
\setlength{\bibhang}{1em}
\setlength{\bibsep}{4pt plus 1pt minus 1pt}
\bibliographystyle{plainnat}
\bibliography{references}

@inproceedings{guo2024skysense,
  title={{SkySense}: A Multi-Modal Remote Sensing Foundation Model Towards Universal Interpretation for Earth Observation Imagery},
  author={Guo, Xin and Lao, Jiangwei and Dang, Bo and Zhang, Yingying and Yu, Lei and Ru, Lixiang and Zhong, Liheng and Huang, Ziyuan and Wu, Kang and Hu, Dingxiang and others},
  booktitle={Proceedings of the IEEE/CVF Conference on Computer Vision and Pattern Recognition},
  pages={27672--27683},
  year={2024}
}

@inproceedings{zhang2025skysensev2,
  title={{SkySense V2}: A Unified Foundation Model for Multi-Modal Remote Sensing},
  author={Zhang, Yingying and Ru, Lixiang and Wu, Kang and Yu, Lei and Liang, Lei and Li, Yansheng and Chen, Jingdong},
  booktitle={Proceedings of the IEEE/CVF International Conference on Computer Vision},
  pages={9136--9146},
  year={2025}
}

@article{wang2024mtp,
  title={{MTP}: Advancing Remote Sensing Foundation Model via Multi-Task Pretraining},
  author={Wang, Di and Zhang, Jing and Xu, Minqiang and Liu, Lin and Wang, Dongsheng and Gao, Erzhong and Han, Chengxi and Guo, Haonan and Du, Bo and Tao, Dacheng and Zhang, Liangpei},
  journal={IEEE Journal of Selected Topics in Applied Earth Observations and Remote Sensing},
  volume={17},
  pages={11632--11654},
  year={2024},
  doi={10.1109/JSTARS.2024.3408154}
}

@article{bi2025ringmoe,
  title={{RingMoE}: Mixture-of-Modality-Experts Multi-Modal Foundation Models for Universal Remote Sensing Image Interpretation},
  author={Bi, Hanbo and Feng, Yingchao and Tong, Boyuan and Wang, Mengyu and Yu, Haichen and Mao, Yongqiang and Chang, Hao and Diao, Wenhui and Wang, Peijin and Yu, Yue and Peng, Hanyang and Zhang, Yehong and Fu, Kun and Sun, Xian},
  journal={IEEE Transactions on Pattern Analysis and Machine Intelligence},
  volume={48},
  number={4},
  pages={4388--4405},
  year={2026},
  doi={10.1109/TPAMI.2025.3643453}
}

@inproceedings{lipman2023flowmatching,
  title={Flow Matching for Generative Modeling},
  author={Lipman, Yaron and Chen, Ricky T. Q. and Ben-Hamu, Heli and Nickel, Maximilian and Le, Matthew},
  booktitle={International Conference on Learning Representations},
  year={2023}
}

@inproceedings{liu2023rectifiedflow,
  title={Flow Straight and Fast: Learning to Generate and Transfer Data with Rectified Flow},
  author={Liu, Xingchao and Gong, Chengyue and Liu, Qiang},
  booktitle={International Conference on Learning Representations},
  year={2023}
}

@inproceedings{sargsyan2026flowdis,
  title={{FlowDIS}: Language-Guided Dichotomous Image Segmentation with Flow Matching},
  author={Sargsyan, Andranik and Navasardyan, Shant},
  booktitle={Proceedings of the IEEE/CVF Conference on Computer Vision and Pattern Recognition},
  pages={42039--42048},
  year={2026}
}

@inproceedings{wang2023painter,
  title={Images Speak in Images: A Generalist Painter for In-Context Visual Learning},
  author={Wang, Xinlong and Wang, Wen and Cao, Yue and Shen, Chunhua and Huang, Tiejun},
  booktitle={Proceedings of the IEEE/CVF Conference on Computer Vision and Pattern Recognition},
  pages={6830--6839},
  year={2023}
}

@inproceedings{geng2024instructdiffusion,
  title={{InstructDiffusion}: A Generalist Modeling Interface for Vision Tasks},
  author={Geng, Zigang and Yang, Binxin and Hang, Tiankai and Li, Chen and Gu, Shuyang and Zhang, Ting and Bao, Jianmin and Zhang, Zheng and Li, Houqiang and Hu, Han and Chen, Dong and Guo, Baining},
  booktitle={Proceedings of the IEEE/CVF Conference on Computer Vision and Pattern Recognition},
  pages={12709--12720},
  year={2024}
}

@inproceedings{gan2024instructcv,
  title={{InstructCV}: Instruction-Tuned Text-to-Image Diffusion Models as Vision Generalists},
  author={Gan, Yulu and Park, Sungwoo and Schubert, Alexander and Philippakis, Anthony and Alaa, Ahmed M.},
  booktitle={International Conference on Learning Representations},
  year={2024}
}

@inproceedings{chen2018gradnorm,
  title={{GradNorm}: Gradient Normalization for Adaptive Loss Balancing in Deep Multitask Networks},
  author={Chen, Zhao and Badrinarayanan, Vijay and Lee, Chen-Yu and Rabinovich, Andrew},
  booktitle={Proceedings of the 35th International Conference on Machine Learning},
  pages={794--803},
  year={2018}
}

@inproceedings{qin2025revisiting,
  title={Revisiting Fairness in Multitask Learning: A Performance-Driven Approach for Variance Reduction},
  author={Qin, Xiaohan and Wang, Xiaoxing and Yan, Junchi},
  booktitle={Proceedings of the IEEE/CVF Conference on Computer Vision and Pattern Recognition},
  pages={20492--20501},
  year={2025},
  doi={10.1109/CVPR52734.2025.01908}
}

@inproceedings{nichol2021improved,
  title={Improved Denoising Diffusion Probabilistic Models},
  author={Nichol, Alexander Quinn and Dhariwal, Prafulla},
  booktitle={Proceedings of the 38th International Conference on Machine Learning},
  pages={8162--8171},
  year={2021}
}

@inproceedings{kim2025adaptive,
  title={Adaptive Non-Uniform Timestep Sampling for Accelerating Diffusion Model Training},
  author={Kim, Myunsoo and Ki, Donghyeon and Shim, Seong-Woong and Lee, Byung-Jun},
  booktitle={Proceedings of the IEEE/CVF Conference on Computer Vision and Pattern Recognition},
  pages={2513--2522},
  year={2025},
  doi={10.1109/CVPR52734.2025.00240}
}

@inproceedings{hu2022lora,
  title={{LoRA}: Low-Rank Adaptation of Large Language Models},
  author={Hu, Edward J. and Shen, Yelong and Wallis, Phillip and Allen-Zhu, Zeyuan and Li, Yuanzhi and Wang, Shean and Wang, Lu and Chen, Weizhu},
  booktitle={International Conference on Learning Representations},
  year={2022}
}

@misc{bfl2026flux2klein,
  author={{Black Forest Labs}},
  title={{FLUX.2 [klein] 4B Base}},
  year={2026},
  note={Model card. Accessed September 22, 2026. Code: \url{https://github.com/black-forest-labs/flux2}},
  url={https://huggingface.co/black-forest-labs/FLUX.2-klein-base-4B}
}

@inproceedings{jakubik2025terramind,
  title={{TerraMind}: Large-Scale Generative Multimodality for Earth Observation},
  author={Jakubik, Johannes and Yang, Felix and Blumenstiel, Benedikt and Scheurer, Erik and Sedona, Rocco and Maurogiovanni, Stefano and Bosmans, Jente and Dionelis, Nikolaos and Marsocci, Valerio and Kopp, Niklas and others},
  booktitle={Proceedings of the IEEE/CVF International Conference on Computer Vision},
  pages={7383--7394},
  year={2025}
}

@article{chen2026any2any,
  title={{Any2Any}: Unified Arbitrary Modality Translation for Remote Sensing},
  author={Chen, Haoyang and Zhang, Jing and Wang, Hebaixu and Wang, Shiqin and Huang, Pohsun and Li, Jiayuan and Guo, Haonan and Wang, Di and Wang, Zheng and Du, Bo},
  journal={arXiv preprint arXiv:2603.04114},
  year={2026},
  doi={10.48550/arXiv.2603.04114}
}

@article{gabeur2026visionbanana,
  title={Image Generators are Generalist Vision Learners},
  author={Gabeur, Valentin and Long, Shangbang and Peng, Songyou and Voigtlaender, Paul and Sun, Shuyang and Bao, Yanan and Truong, Karen and Wang, Zhicheng and Zhou, Wenlei and Barron, Jonathan T. and others},
  journal={arXiv preprint arXiv:2604.20329},
  year={2026},
  doi={10.48550/arXiv.2604.20329}
}

@article{qwen2025qwen3,
  title={{Qwen3} Technical Report},
  author={{Qwen Team}},
  journal={arXiv preprint arXiv:2505.09388},
  year={2025},
  doi={10.48550/arXiv.2505.09388}
}

@article{ji2019whu,
  title={Fully Convolutional Networks for Multisource Building Extraction From an Open Aerial and Satellite Imagery Data Set},
  author={Ji, Shunping and Wei, Shiqing and Lu, Meng},
  journal={IEEE Transactions on Geoscience and Remote Sensing},
  volume={57},
  number={1},
  pages={574--586},
  year={2019}
}

@article{shi2021dmbc,
  title={A Multi-Task Network With Distance--Mask--Boundary Consistency Constraints for Building Extraction From Aerial Images},
  author={Shi, Furong and Zhang, Tong},
  journal={Remote Sensing},
  volume={13},
  number={14},
  pages={2656},
  year={2021},
  doi={10.3390/rs13142656}
}

@inproceedings{shermeyer2020spacenet6,
  title={{SpaceNet 6}: Multi-Sensor All Weather Mapping Dataset},
  author={Shermeyer, Jacob and Hogan, Daniel and Brown, Jason and Van Etten, Adam and Weir, Nicholas and Pacifici, Fabio and H{\"a}nsch, Ronny and Bastidas, Alexei and Soenen, Scott and Bacastow, Todd and Lewis, Ryan},
  booktitle={Proceedings of the IEEE/CVF Conference on Computer Vision and Pattern Recognition Workshops},
  pages={768--777},
  year={2020},
  doi={10.1109/CVPRW50498.2020.00106}
}

@article{shen2021s2looking,
  title={{S2Looking}: A Satellite Side-Looking Dataset for Building Change Detection},
  author={Shen, Li and Lu, Yao and Chen, Hao and Wei, Hao and Xie, Donghai and Yue, Jiabao and Chen, Rui and Lv, Shouye and Jiang, Bitao},
  journal={Remote Sensing},
  volume={13},
  number={24},
  pages={5094},
  year={2021},
  doi={10.3390/rs13245094}
}

@inproceedings{zhao2024urbansarfloods,
  title={{UrbanSARFloods}: {Sentinel-1 SLC}-Based Benchmark Dataset for Urban and Open-Area Flood Mapping},
  author={Zhao, Jie and Xiong, Zhitong and Zhu, Xiao Xiang},
  booktitle={Proceedings of the IEEE/CVF Conference on Computer Vision and Pattern Recognition Workshops},
  pages={419--429},
  year={2024}
}

@inproceedings{wang2021loveda,
  title={{LoveDA}: A Remote Sensing Land-Cover Dataset for Domain Adaptive Semantic Segmentation},
  author={Wang, Junjue and Zheng, Zhuo and Ma, Ailong and Lu, Xiaoyan and Zhong, Yanfei},
  booktitle={Proceedings of the Neural Information Processing Systems Track on Datasets and Benchmarks},
  volume={1},
  year={2021}
}

@inproceedings{xia2023openearthmap,
  title={{OpenEarthMap}: A Benchmark Dataset for Global High-Resolution Land Cover Mapping},
  author={Xia, Junshi and Yokoya, Naoto and Adriano, Bruno and Broni-Bediako, Clifford},
  booktitle={Proceedings of the IEEE/CVF Winter Conference on Applications of Computer Vision},
  pages={6254--6264},
  year={2023}
}

@article{xia2025openearthmapsar,
  title={{OpenEarthMap-SAR}: A Benchmark Synthetic Aperture Radar Dataset for Global High-Resolution Land Cover Mapping [Software and Data Sets]},
  author={Xia, Junshi and Chen, Hongruixuan and Broni-Bediako, Clifford and Wei, Yimin and Song, Jian and Yokoya, Naoto},
  journal={IEEE Geoscience and Remote Sensing Magazine},
  volume={13},
  number={4},
  year={2025},
  pages={476--487},
  doi={10.1109/MGRS.2025.3599512}
}

@inproceedings{rottensteiner2012isprs,
  title={The {ISPRS} Benchmark on Urban Object Classification and {3D} Building Reconstruction},
  author={Rottensteiner, F. and Sohn, G. and Jung, J. and Gerke, M. and Baillard, C. and Benitez, S. and Breitkopf, U.},
  booktitle={ISPRS Annals of the Photogrammetry, Remote Sensing and Spatial Information Sciences},
  volume={I-3},
  pages={293--298},
  year={2012},
  doi={10.5194/isprsannals-I-3-293-2012}
}

@article{liu2024dfc23,
  title={Large-Scale Fine-Grained Building Classification and Height Estimation for Semantic Urban Reconstruction: Outcome of the 2023 {IEEE GRSS} Data Fusion Contest},
  author={Liu, Guozhang and Peng, Baochai and Liu, Ting and Zhang, Pan and Yuan, Mengke and Lu, Chaoran and Cao, Ningning and Zhang, Sen and Huang, Simin and Wang, Tao and others},
  journal={IEEE Journal of Selected Topics in Applied Earth Observations and Remote Sensing},
  volume={17},
  pages={11194--11207},
  year={2024}
}

@inproceedings{liu2024rmsin,
  title={Rotated Multi-Scale Interaction Network for Referring Remote Sensing Image Segmentation},
  author={Liu, Sihan and Ma, Yiwei and Zhang, Xiaoqing and Wang, Haowei and Ji, Jiayi and Sun, Xiaoshuai and Ji, Rongrong},
  booktitle={Proceedings of the IEEE/CVF Conference on Computer Vision and Pattern Recognition},
  pages={26658--26668},
  year={2024}
}

@inproceedings{xin2026segearthr2,
  title={{SegEarth-R2}: Towards Comprehensive Language-Guided Segmentation for Remote Sensing Images},
  author={Xin, Zepeng and Li, Kaiyu and Chen, Luodi and Li, Wanchen and Xiao, Yuchen and Qiao, Hui and Zhang, Weizhan and Meng, Deyu and Cao, Xiangyong},
  booktitle={Proceedings of the IEEE/CVF Conference on Computer Vision and Pattern Recognition},
  pages={13199--13210},
  year={2026}
}

@article{cheng2022aopg,
  title={Anchor-Free Oriented Proposal Generator for Object Detection},
  author={Cheng, Gong and Wang, Jiabao and Li, Ke and Xie, Xingxing and Lang, Chunbo and Yao, Yanqing and Han, Junwei},
  journal={IEEE Transactions on Geoscience and Remote Sensing},
  volume={60},
  pages={1--11},
  year={2022}
}

@inproceedings{zhang2026unichange,
  title={{UniChange}: Unifying Change Detection with Multimodal Large Language Model},
  author={Zhang, Xu and Li, Danyang and Dong, Xiaohang and Wu, Tianhao and Yu, Hualong and Wang, Jianye and Li, Qicheng and Li, Xiang},
  booktitle={Proceedings of the IEEE/CVF Conference on Computer Vision and Pattern Recognition},
  pages={42169--42179},
  year={2026}
}

@article{wang2024metasegnet,
  title={{MetaSegNet}: Metadata-Collaborative Vision-Language Representation Learning for Semantic Segmentation of Remote Sensing Images},
  author={Wang, Libo and Dong, Sijun and Chen, Ying and Meng, Xiaoliang and Fang, Shenghui and Fei, Songlin},
  journal={IEEE Transactions on Geoscience and Remote Sensing},
  volume={62},
  pages={1--11},
  year={2024},
  doi={10.1109/TGRS.2024.3477548}
}

@inproceedings{zhou2026tricdnet,
  title={{TriCDNet}: A Multi-Scale Tri-Stream Interaction Network for Building Change Detection},
  author={Zhou, Renjie},
  booktitle={The International Archives of the Photogrammetry, Remote Sensing and Spatial Information Sciences},
  volume={XLVIII-3/W4-2025},
  pages={63--69},
  year={2026},
  doi={10.5194/isprs-archives-XLVIII-3-W4-2025-63-2026}
}

@article{dong2024changeclip,
  title={{ChangeCLIP}: Remote Sensing Change Detection with Multimodal Vision-Language Representation Learning},
  author={Dong, Sijun and Wang, Libo and Du, Bo and Meng, Xiaoliang},
  journal={ISPRS Journal of Photogrammetry and Remote Sensing},
  volume={208},
  pages={53--69},
  year={2024},
  doi={10.1016/j.isprsjprs.2024.01.004}
}

@inproceedings{cheng2022mask2former,
  title={Masked-Attention Mask Transformer for Universal Image Segmentation},
  author={Cheng, Bowen and Misra, Ishan and Schwing, Alexander G. and Kirillov, Alexander and Girdhar, Rohit},
  booktitle={Proceedings of the IEEE/CVF Conference on Computer Vision and Pattern Recognition},
  pages={1290--1299},
  year={2022}
}

@inproceedings{zou2025d2ls,
  title={Dynamic Dictionary Learning for Remote Sensing Image Segmentation},
  author={Zou, Xuechao and Li, Yue and Zhang, Shun and Li, Kai and Wang, Shiying and Tao, Pin and Xing, Junliang and Lang, Congyan},
  booktitle={Proceedings of the IEEE/CVF International Conference on Computer Vision},
  pages={22457--22466},
  year={2025}
}

@article{he2024uanet,
  title={{UANet}: An Uncertainty-Aware Network for Building Extraction From Remote Sensing Images},
  author={Li, Jiepan and He, Wei and Cao, Weinan and Zhang, Liangpei and Zhang, Hongyan},
  journal={IEEE Transactions on Geoscience and Remote Sensing},
  volume={62},
  pages={1--13},
  year={2024},
  doi={10.1109/TGRS.2024.3361211}
}

@inproceedings{zhao2025ccs,
  title={Convex Combination Star Shape Prior for Data-Driven Image Semantic Segmentation},
  author={Zhao, Xinyu and Xie, Jun and Chen, Shengzhe and Liu, Jun},
  booktitle={Proceedings of the IEEE/CVF Conference on Computer Vision and Pattern Recognition},
  pages={14068--14077},
  year={2025}
}

@article{wang2024bidnet,
  title={Extracting Building Footprints in {SAR} Images via Distilling Boundary Information From Optical Images},
  author={Wang, Yuxuan and Zeng, Lanxin and Yang, Wen and Kang, Jian and Yu, Huai and Datcu, Mihai and Xia, Gui-Song},
  journal={IEEE Transactions on Geoscience and Remote Sensing},
  volume={62},
  pages={1--15},
  year={2024},
  doi={10.1109/TGRS.2024.3359704}
}

@article{pei2026aemkd,
  title={{AEM-KD}: Adversarial Entropy Minimization for Cross-Modality Knowledge Distillation in {SAR} Building Segmentation},
  author={Pei, Hao and Ni, Peishuang and Qiao, Yiguo and Chen, Jianlai and Liu, Wenkang and Yu, Hanwen and Wu, Fan and Xu, Gang and Xing, Mengdao},
  journal={IEEE Journal of Selected Topics in Applied Earth Observations and Remote Sensing},
  volume={19},
  pages={16201--16219},
  year={2026},
  doi={10.1109/JSTARS.2026.3686377}
}

@article{wei2026cmhcl,
  title={Optical and {SAR} Cross-Modal Hallucination Collaborative Learning for Remote Sensing Missing-Modality Building Footprint Extraction},
  author={Wei, Tianyu and Chen, He and Liu, Wenchao and Chen, Liang and Gu, Panzhe and Wang, Jue},
  journal={IEEE Journal of Selected Topics in Applied Earth Observations and Remote Sensing},
  volume={19},
  pages={1183--1196},
  year={2026},
  doi={10.1109/JSTARS.2025.3638382}
}

@misc{robinson2024chesapeake,
  title={Seeing the Roads Through the Trees: A Benchmark for Modeling Spatial Dependencies with Aerial Imagery},
  author={Robinson, Caleb and Corley, Isaac and Ortiz, Anthony and Dodhia, Rahul and Lavista Ferres, Juan M. and Najafirad, Peyman},
  year={2024},
  eprint={2401.06762},
  archivePrefix={arXiv},
  primaryClass={cs.CV},
  url={https://arxiv.org/abs/2401.06762}
}

@article{lv2026s5,
  title={{S5}: Scalable Semi-Supervised Semantic Segmentation in Remote Sensing},
  author={Lv, Liang and Wang, Di and Zhang, Jing and Zhang, Lefei},
  journal={Proceedings of the AAAI Conference on Artificial Intelligence},
  volume={40},
  number={10},
  pages={7726--7734},
  year={2026},
  doi={10.1609/aaai.v40i10.37715}
}

@inproceedings{wang2025selectivemae,
  title={Harnessing Massive Satellite Imagery with Efficient Masked Image Modeling},
  author={Wang, Fengxiang and Wang, Hongzhen and Wang, Di and Guo, Zonghao and Zhong, Zhenyu and Lan, Long and Yang, Wenjing and Zhang, Jing},
  booktitle={Proceedings of the IEEE/CVF International Conference on Computer Vision},
  pages={6935--6947},
  year={2025}
}

@inproceedings{tan2025fedsense,
  title={Towards Privacy-Preserved Pre-Training of Remote Sensing Foundation Models with Federated Mutual-Guidance Learning},
  author={Tan, Jieyi and Zhang, Chengwei and Dang, Bo and Li, Yansheng},
  booktitle={Proceedings of the IEEE/CVF International Conference on Computer Vision},
  pages={1804--1814},
  year={2025},
  doi={10.1109/ICCV51701.2025.00176}
}

@article{chen2024heightformer,
  title={{HeightFormer}: A Multilevel Interaction and Image-Adaptive Classification--Regression Network for Monocular Height Estimation with Aerial Images},
  author={Chen, Zhan and Zhang, Yidan and Qi, Xiyu and Mao, Yongqiang and Zhou, Xin and Wang, Lei and Ge, Yunping},
  journal={Remote Sensing},
  volume={16},
  number={2},
  pages={295},
  year={2024},
  doi={10.3390/rs16020295}
}

@article{lei2025raephrnet,
  title={A Relation-Aware and Edge-Preserving Height Refinement Network for Single-View Height Estimation From Remote Sensing Images},
  author={Lei, Yang and Jiang, Wanshou},
  journal={IEEE Journal of Selected Topics in Applied Earth Observations and Remote Sensing},
  volume={18},
  pages={12988--13002},
  year={2025},
  doi={10.1109/JSTARS.2025.3567064}
}

@inproceedings{mao2023light,
  title={{LIGHT}: Joint Individual Building Extraction and Height Estimation from Satellite Images through a Unified Multitask Learning Network},
  author={Mao, Yongqiang and Sun, Xian and Huang, Xingliang and Chen, Kaiqiang},
  booktitle={IEEE International Geoscience and Remote Sensing Symposium},
  year={2023},
  url={https://arxiv.org/abs/2304.01090}
}

@article{yang2026mars,
  title={{MaRS}: A Multi-Modality Very-High-Resolution Remote Sensing Foundation Model with Cross-Granularity Meta-Modality Learning},
  author={Yang, Ruoyu and Liu, Yinhe and Yan, Heng and Zhou, Yiheng and Fu, Yihan and Luo, Han and Zhong, Yanfei},
  journal={Proceedings of the AAAI Conference on Artificial Intelligence},
  volume={40},
  number={14},
  pages={11685--11693},
  year={2026},
  doi={10.1609/aaai.v40i14.38153}
}

@article{yang2026changerwkv,
  title={Beyond Quadratic: Linear-Time Change Detection with {RWKV}},
  author={Yang, Zhenyu and Pei, Gensheng and Chen, Tao and Yuan, Xia and Zhang, Haofeng and Shu, Xiangbo and Yao, Yazhou},
  journal={Proceedings of the AAAI Conference on Artificial Intelligence},
  volume={40},
  number={14},
  pages={11811--11819},
  year={2026},
  doi={10.1609/aaai.v40i14.38167}
}

@article{rong2026rs2sam2,
  title={{RS2-SAM2}: Customized {SAM2} for Referring Remote Sensing Image Segmentation},
  author={Rong, Fu and Lan, Meng and Zhang, Qian and Zhang, Lefei},
  journal={Proceedings of the AAAI Conference on Artificial Intelligence},
  volume={40},
  number={11},
  pages={8751--8759},
  year={2026},
  doi={10.1609/aaai.v40i11.37828}
}

@inproceedings{liu2026visionreasoner,
  title={{VisionReasoner}: Unified Reasoning-Integrated Visual Perception via Reinforcement Learning},
  author={Liu, Yuqi and Qu, Tianyuan and Zhong, Zhisheng and Peng, Bohao and Liu, Shu and Yu, Bei and Jia, Jiaya},
  booktitle={International Conference on Learning Representations},
  year={2026},
  url={https://openreview.net/forum?id=QoDOwjsbAq}
}

@inproceedings{lai2024lisa,
  title={{LISA}: Reasoning Segmentation via Large Language Model},
  author={Lai, Xin and Tian, Zhuotao and Chen, Yukang and Li, Yanwei and Yuan, Yuhui and Liu, Shu and Jia, Jiaya},
  booktitle={Proceedings of the IEEE/CVF Conference on Computer Vision and Pattern Recognition},
  pages={9579--9589},
  year={2024}
}

@inproceedings{shabbir2025geopixel,
  title={{GeoPixel}: Pixel Grounding Large Multimodal Model in Remote Sensing},
  author={Shabbir, Akashah and Zumri, Mohammed and Bennamoun, Mohammed and Khan, Fahad Shahbaz and Khan, Salman},
  booktitle={Proceedings of the 42nd International Conference on Machine Learning},
  series={Proceedings of Machine Learning Research},
  volume={267},
  pages={54095--54111},
  year={2025}
}

@article{yao2026uaglnet,
  title={{UAGLNet}: Uncertainty-Aggregated Global-Local Fusion Network with Cooperative CNN-Transformer for Building Extraction},
  author={Yao, Siyuan and Liu, Dongxiu and Li, Taotao and Li, Shengjie and Ren, Wenqi and Cao, Xiaochun},
  journal={IEEE Transactions on Geoscience and Remote Sensing},
  volume={64},
  pages={1--14},
  year={2026},
  doi={10.1109/TGRS.2025.3645320}
}

@inproceedings{huang2024mutdet,
  title={{MutDet}: Mutually Optimizing Pre-training for Remote Sensing Object Detection},
  author={Huang, Ziyue and Feng, Yongchao and Liu, Qingjie and Wang, Yunhong},
  booktitle={Computer Vision -- ECCV 2024},
  pages={1--17},
  year={2024},
  doi={10.1007/978-3-031-73254-6_1}
}

@article{li2025uabcd,
  title={Overcoming the Uncertainty Challenges in Detecting Building Changes from Remote Sensing Images},
  author={Li, Jiepan and He, Wei and Li, Zhuohong and Guo, Yujun and Zhang, Hongyan},
  journal={ISPRS Journal of Photogrammetry and Remote Sensing},
  volume={220},
  pages={1--17},
  year={2025},
  doi={10.1016/j.isprsjprs.2024.11.017}
}

@inproceedings{zhang2025corrclip,
  title={{CorrCLIP}: Reconstructing Patch Correlations in {CLIP} for Open-Vocabulary Semantic Segmentation},
  author={Zhang, Dengke and Liu, Fagui and Tang, Quan},
  booktitle={Proceedings of the IEEE/CVF International Conference on Computer Vision},
  pages={24677--24687},
  year={2025}
}

@inproceedings{li2025segearthov,
  title={{SegEarth-OV}: Towards Training-Free Open-Vocabulary Segmentation for Remote Sensing Images},
  author={Li, Kaiyu and Liu, Ruixun and Cao, Xiangyong and Bai, Xueru and Zhou, Feng and Meng, Deyu and Wang, Zhi},
  booktitle={Proceedings of the IEEE/CVF Conference on Computer Vision and Pattern Recognition},
  pages={10545--10556},
  year={2025},
  doi={10.1109/CVPR52734.2025.00986}
}

@inproceedings{lee2025hausdorff,
  title={Hausdorff Distance Matching with Adaptive Query Denoising for Rotated Detection Transformer},
  author={Lee, Hakjin and Song, Minki and Koo, Jamyoung and Seo, Junghoon},
  booktitle={Proceedings of the IEEE/CVF Winter Conference on Applications of Computer Vision},
  pages={1872--1882},
  year={2025}
}

@article{du2024lumnet,
  title={{LUMNet}: Land Use Knowledge Guided Multiscale Network for Height Estimation From Single Remote Sensing Images},
  author={Du, Shouhang and Xing, Jianghe and Wang, Shaoyu and Xiao, Xiongwu and Li, Jun and Liu, Hao},
  journal={IEEE Geoscience and Remote Sensing Letters},
  volume={21},
  pages={1--5},
  year={2024},
  doi={10.1109/LGRS.2024.3374526}
}

@inproceedings{jang2025mmr,
  title={{MMR}: A Large-Scale Benchmark Dataset for Multi-Target and Multi-Granularity Reasoning Segmentation},
  author={Jang, Donggon and Cho, Yucheol and Lee, Suin and Kim, Taehyeon and Kim, Daeshik},
  booktitle={The Thirteenth International Conference on Learning Representations},
  year={2025}
}

@article{xiong2024dofa,
  title={Neural Plasticity-Inspired Multimodal Foundation Model for Earth Observation},
  author={Xiong, Zhitong and Wang, Yi and Zhang, Fahong and Stewart, Adam J. and Hanna, Jo{\"e}lle and Borth, Damian and Papoutsis, Ioannis and Le Saux, Bertrand and Camps-Valls, Gustau and Zhu, Xiao Xiang},
  journal={arXiv preprint arXiv:2403.15356},
  year={2024},
  doi={10.48550/arXiv.2403.15356}
}

@misc{isprs_potsdam,
  author={{ISPRS}},
  title={{2D} Semantic Labeling Contest -- {Potsdam}},
  year={n.d.},
  url={https://www.isprs.org/resources/datasets/benchmarks/UrbanSemLab/2d-sem-label-potsdam.aspx},
  note={Dataset resource. Accessed September 22, 2026}
}

@article{disoptnet,
  title   = {{DisOptNet}: Distilling Semantic Knowledge From Optical Images for Weather-Independent Building Segmentation},
  author  = {Kang, Jian and Wang, Zhirui and Zhu, Ruoxin and Xia, Junshi and Sun, Xian and Fernandez-Beltran, Ruben and Plaza, Antonio},
  journal = {IEEE Transactions on Geoscience and Remote Sensing},
  volume  = {60},
  pages   = {1--15},
  year    = {2022},
  doi     = {10.1109/TGRS.2022.3165209}
}

@article{htcdc,
  title   = {{HTC-DC Net}: Monocular Height Estimation From Single Remote Sensing Images},
  author  = {Chen, Sining and Shi, Yilei and Xiong, Zhitong and Zhu, Xiao Xiang},
  journal = {IEEE Transactions on Geoscience and Remote Sensing},
  volume  = {61},
  pages   = {1--18},
  year    = {2023},
  doi     = {10.1109/TGRS.2023.3321255}
}

@article{chen2021stt,
  title={Building Extraction from Remote Sensing Images with Sparse Token Transformers},
  author={Chen, Keyan and Zou, Zhengxia and Shi, Zhenwei},
  journal={Remote Sensing},
  volume={13},
  number={21},
  pages={4441},
  year={2021},
  doi={10.3390/rs13214441}
}

@inproceedings{chen2018deeplabv3plus,
  title={Encoder-Decoder with Atrous Separable Convolution for Semantic Image Segmentation},
  author={Chen, Liang-Chieh and Zhu, Yukun and Papandreou, George and Schroff, Florian and Adam, Hartwig},
  booktitle={Proceedings of the European Conference on Computer Vision},
  pages={801--818},
  year={2018},
  url={https://www.ecva.net/papers/eccv_2018/papers_ECCV/html/Liang-Chieh_Chen_Encoder-Decoder_with_Atrous_ECCV_2018_paper.php}
}

@inproceedings{song2024synrs3d,
  title={{SynRS3D}: A Synthetic Dataset for Global {3D} Semantic Understanding from Monocular Remote Sensing Imagery},
  author={Song, Jian and Chen, Hongruixuan and Xuan, Weihao and Xia, Junshi and Yokoya, Naoto},
  booktitle={Advances in Neural Information Processing Systems},
  volume={37},
  year={2024},
  url={https://proceedings.neurips.cc/paper_files/paper/2024/hash/d4f3da38b491b44d40a0d5a5b37134ba-Abstract-Datasets_and_Benchmarks_Track.html}
}

@article{zhang2024asanet,
  title={{ASANet}: Asymmetric Semantic Aligning Network for {RGB} and {SAR} Image Land Cover Classification},
  author={Zhang, Pan and Peng, Baochai and Lu, Chaoran and Huang, Quanjin and Liu, Dongsheng},
  journal={ISPRS Journal of Photogrammetry and Remote Sensing},
  volume={218},
  pages={574--587},
  year={2024},
  doi={10.1016/j.isprsjprs.2024.09.025}
}

@inproceedings{wei2026mmovseg,
  title={{MM-OVSeg}: Multimodal Optical--{SAR} Fusion for Open-Vocabulary Segmentation in Remote Sensing},
  author={Wei, Yimin and Xiao, Aoran and Chen, Hongruixuan and Xia, Junshi and Yokoya, Naoto},
  booktitle={Proceedings of the IEEE/CVF Conference on Computer Vision and Pattern Recognition},
  pages={42202--42212},
  year={2026},
  url={https://github.com/Jimmyxichen/MM-OVSeg}
}

@inproceedings{hoyer2022daformer,
  title={{DAFormer}: Improving Network Architectures and Training Strategies for Domain-Adaptive Semantic Segmentation},
  author={Hoyer, Lukas and Dai, Dengxin and Van Gool, Luc},
  booktitle={Proceedings of the IEEE/CVF Conference on Computer Vision and Pattern Recognition},
  pages={9924--9935},
  year={2022},
  url={https://openaccess.thecvf.com/content/CVPR2022/html/Hoyer_DAFormer_Improving_Network_Architectures_and_Training_Strategies_for_Domain-Adaptive_Semantic_CVPR_2022_paper.html}
}

@inproceedings{peng2026roadgie,
  title={{RoadGIE}: Towards a Global-Scale Aerial Benchmark for Generalizable Interactive Road Extraction},
  author={Peng, Chenxu and Wang, Chenxu and Dai, Yimian and Liu, Yongxiang and Cheng, Ming-Ming and Li, Xiang},
  booktitle={Proceedings of the IEEE/CVF Conference on Computer Vision and Pattern Recognition},
  pages={13285--13295},
  year={2026},
  url={https://github.com/chaineypung/RoadGIE}
}

@article{zhao2024orientedformer,
  title={{OrientedFormer}: An End-to-End Transformer-Based Oriented Object Detector in Remote Sensing Images},
  author={Zhao, Jiaqi and Ding, Zeyu and Zhou, Yong and Zhu, Hancheng and Du, Wen-Liang and Yao, Rui and El Saddik, Abdulmotaleb},
  journal={IEEE Transactions on Geoscience and Remote Sensing},
  volume={62},
  pages={1--16},
  year={2024},
  doi={10.1109/TGRS.2024.3456240}
}
\endgroup

\clearpage
\appendix
\raggedbottom
\setlength{\textfloatsep}{10pt plus 2pt minus 2pt}
\setlength{\intextsep}{8pt plus 2pt minus 2pt}
\clubpenalty=10000
\widowpenalty=10000
\displaywidowpenalty=10000
\makeatletter
\setlength{\@fptop}{0pt}
\setlength{\@fpsep}{14pt plus 2pt minus 2pt}
\makeatother

\section*{Appendix}

\section{Dataset Details and Evaluation Protocols}
\label{app:dataset_details}

We evaluate six tasks across thirteen public remote-sensing datasets, spanning
optical and SAR imagery, single- and bi-temporal inputs, continuous prediction,
language-conditioned segmentation, and instance-level detection. We follow the
official or released dataset splits and evaluation protocols whenever available.
Target-codec choices are determined using the validation split, or the training
split when no validation split is provided. Dataset-specific preprocessing and
evaluation details are described below.

\paragraph{Dataset-specific protocols.}
For building extraction, WHU Building uses RGB aerial imagery and follows the
provider-defined folders~\citep{ji2019whu}. For SpaceNet~6, we use SAR imagery and follow the data partitioning strategy commonly adopted by recent baselines. The released building polygons are rasterized onto the corresponding image grids and converted into binary targets using the target codec~\citep{shermeyer2020spacenet6}.

For change detection, WHU-CD and LEVIR-CD+ use bi-temporal RGB imagery.
WHU-CD follows the released $512\times512$ training and test partitions,
containing 1,260 and 690 bi-temporal image pairs, respectively, while
LEVIR-CD+ follows its released train/test partition~\citep{shen2021s2looking}.
UrbanSARFloods uses bi-temporal SAR observations and follows the released
training/validation lists and held-out event test cases~\citep{zhao2024urbansarfloods}.
Geometric transforms are synchronized across both temporal observations and
the target.

For semantic segmentation, LoveDA and OpenEarthMap use RGB imagery, whereas
OpenEarthMap-SAR uses SAR imagery; all three datasets follow their released
partitions~\citep{wang2021loveda,xia2023openearthmap,xia2025openearthmapsar}.
Dataset-native labels are mapped to the canonical visual codebook using
deterministic lookup tables, and ignored or unlabeled pixels are excluded from
both training loss and evaluation.

For height estimation, Potsdam uses RGB imagery with the 24/14
labeled-train/held-out-test split~\citep{isprs_potsdam}. DFC23 Track~2 provides both RGB and SAR observations and follows the
commonly used data split adopted by prior baseline methods~\citep{liu2024dfc23}.
Both modalities are used as height-estimation samples during joint training,
while the reported benchmark result uses RGB imagery at inference. Invalid height pixels are masked before the
log-compressed target encoding.

For referring segmentation, RRSIS-D and LaSeRS use RGB imagery paired with
text expressions. RRSIS-D uses its official split~\citep{liu2024rmsin},
whereas LaSeRS uses the released train/test annotations without
repartitioning~\citep{xin2026segearthr2}. For rotated detection, DIOR-R uses
RGB imagery and retains its official train/validation/test
split~\citep{cheng2022aopg}.

\paragraph{Common preprocessing.}
Intensity-normalization statistics are estimated from the training partition.
Geometric augmentations are synchronized across inputs and targets, including
both temporal observations for change detection. Evaluation predictions are
restored to the native image grid before applying the deterministic recovery
operator $R_d$.

\subsection{Text Prompts and Language Conditioning}
\label{app:text_prompts}

The language condition is constructed as $c_T=T(q)$. For building extraction,
change detection, semantic segmentation, height estimation, and rotated
detection, $q$ is a fixed instruction determined by the dataset and input
modality. Referring segmentation instead uses the sample-specific expression
provided by the dataset as $q$. Figures~\ref{fig:text_prompts_general}
and~\ref{fig:text_prompts_semantic} list the configured fixed instructions.
The same text-conditioning protocol is used for training and evaluation.

\paragraph{Prompt assignment and referring expressions.}
WHU Building uses the optical building prompt, and SpaceNet~6 SAR uses the SAR
building prompt. WHU-CD and LEVIR-CD+ share the optical change prompt, whereas
UrbanSARFloods uses the SAR flood-change prompt. Potsdam and the optical branch
of DFC23 Track~2 share the optical height prompt, while the SAR branch uses its
SAR counterpart. The semantic prompts distinguish LoveDA, OpenEarthMap
optical, and OpenEarthMap-SAR.

\begin{figure}[H]
\noindent\textbf{Task instructions:}\par\vspace{3pt}
\begin{promptbox}
\promptentry{Building / optical}{\detokenize{Generate a binary building-footprint map for the optical aerial image. Buildings are white; background is black.}}
\promptdivider
\promptentry{Building / SAR}{\detokenize{Generate a binary building-footprint map for the synthetic aperture radar image. Buildings are white; background is black.}}
\promptdivider
\promptentry{Change / optical pair}{\detokenize{Generate a binary building-change map for the before and after optical aerial images. Changed buildings are white; unchanged areas are black.}}
\promptdivider
\promptentry{Change / SAR pair}{\detokenize{Generate a per-pixel flood-change map for the pre-event and co-event synthetic aperture radar images, using this RGB color mapping: non-flood <0,0,0>; open-area flood <0,0,255>; urban flood <255,0,0>.}}
\promptdivider
\promptentry{Height / optical}{\detokenize{Generate a dense height map for the optical aerial image. Encode log1p height from 0 to 200 meters as grayscale, from black to white.}}
\promptdivider
\promptentry{Height / SAR}{\detokenize{Generate a dense height map for the synthetic aperture radar image. Encode log1p height from 0 to 200 meters as grayscale, from black to white.}}
\promptdivider
\promptentry{Detection / DIOR-R}{\detokenize{Generate a class-coded mask of all DIOR-R object classes.}}
\end{promptbox}
\caption{Fixed text prompts for building extraction, change detection, height
estimation, and rotated detection. Prompts are shown verbatim.}
\label{fig:text_prompts_general}
\end{figure}

\begin{figure}[H]
\noindent\textbf{Semantic-segmentation instructions:}\par\vspace{3pt}
\begin{promptbox}
\promptentry{LoveDA / optical}{\detokenize{Generate a per-pixel LoveDA semantic map for the optical aerial image, using the VAE-aware canonical palette: background <0,0,0>; building <222,31,7>; road <255,255,255>; water <0,69,255>; barren <255,0,255>; forest <34,97,38>; agriculture <0,255,255>.}}
\promptdivider
\promptentry{OpenEarthMap / optical}{Generate an OpenEarthMap optical semantic map using the VAE-aware canonical palette. RGB colors: background black \#000000; bareland magenta \#FF00FF; rangeland bright green \#00FF24; developed space gray \#949494; road white \#FFFFFF; tree dark green \#226126; water blue \#0045FF; agriculture land cyan \#00FFFF; building red \#DE1F07.}
\promptdivider
\promptentry{OpenEarthMap-SAR / SAR}{Generate an OpenEarthMap SAR semantic map using the VAE-aware canonical palette. RGB colors: background black \#000000; bareland magenta \#FF00FF; rangeland bright green \#00FF24; developed space gray \#949494; road white \#FFFFFF; tree dark green \#226126; water blue \#0045FF; agriculture land cyan \#00FFFF; building red \#DE1F07.}
\end{promptbox}
\caption{Dataset- and modality-specific semantic-segmentation prompts.
RGB triplets and hexadecimal colors are explicitly included in the prompts.}
\label{fig:text_prompts_semantic}
\end{figure}

For RRSIS-D, we use the annotation's \texttt{raw} field, falling back to
\texttt{sent} when unavailable; for LaSeRS, we use the \texttt{description}
field. The dataset expressions are otherwise used unchanged except for
whitespace trimming.

\Needspace{8\baselineskip}
\paragraph{Text encoding.}
We use the frozen Qwen3 text conditioner associated with FLUX.2-Klein Base.
Each prompt is passed as the content of a single user message through the
tokenizer's chat template, with
\path{add_generation_prompt=True} and
\path{enable_thinking=False}. The resulting sequence is padded or truncated
to 128 tokens, including the chat-template tokens. Hidden states at indices
9, 18, and 27 are concatenated along the feature dimension to obtain a
$128\times7680$ condition. Fixed instructions are encoded once and cached,
while referring conditions are cached per expression.

\subsection{Evaluation Metrics}
\label{app:evaluation_metrics}

Table~\ref{tab:metric_assignment} lists the evaluation measure used for every
dataset. Unless stated otherwise, overlap and accuracy scores are reported as
percentages, and ignored or invalid pixels are excluded according to the
official evaluation protocol.

\begin{table}[H]
\caption{Evaluation metrics for the thirteen datasets. Arrows indicate the preferred direction.}
\label{tab:metric_assignment}
\centering
\tablebodyfont
\setlength{\tabcolsep}{5pt}
\renewcommand{\arraystretch}{1.06}
\begin{tabular}{lll}
\toprule
Task & Dataset & Reported metric(s) \\
\midrule
Building extraction & WHU Building & IoU $\uparrow$ \\
                    & SpaceNet 6 (SAR) & IoU $\uparrow$ \\
Change detection    & WHU-CD & IoU $\uparrow$ \\
                    & LEVIR-CD+ & IoU $\uparrow$ \\
                    & UrbanSARFloods (SAR) & mIoU $\uparrow$ \\
Semantic segmentation & LoveDA & mIoU $\uparrow$ \\
                    & OpenEarthMap-RGB & mIoU $\uparrow$ \\
                    & OpenEarthMap-SAR & mIoU $\uparrow$ \\
Height estimation  & Potsdam & RMSE (m) $\downarrow$ \\
                    & DFC23 Track 2 & $\delta_1$ $\uparrow$ \\
Referring segmentation & RRSIS-D & cIoU / gIoU $\uparrow$ \\
                    & LaSeRS & cIoU / gIoU $\uparrow$ \\
Rotated detection  & DIOR-R & mAP50 $\uparrow$ \\
\bottomrule
\end{tabular}
\end{table}

\paragraph{Unified Task Score (UTS).}
For controlled comparisons among unified variants, let
$M_d^{(a)}\in[0,1]$ be the selected higher-is-better score for method $a$ on
dataset $d$. We first average the UTS-eligible datasets within each task and
then assign equal weight to all six tasks:
\begin{equation}
S_\tau^{(a)}=
\frac{100}{|\Dset_\tau^{\mathrm{UTS}}|}
\sum_{d\in\Dset_\tau^{\mathrm{UTS}}}M_d^{(a)},
\qquad
\uts(a)=\frac{1}{6}\sum_{\tau\in\Tset}S_\tau^{(a)},
\quad |\Tset|=6.
\label{eq:uts}
\end{equation}

We use IoU for building extraction, IoU or mIoU for change detection, mIoU for
semantic segmentation, $\delta_1$ for height estimation, gIoU for referring
segmentation, and mAP50 for rotated detection. Potsdam RMSE and the
complementary cIoU results are reported separately and excluded from UTS.
UTS is reported for controlled comparisons among unified-model variants,
alongside the native benchmark metrics.

\paragraph{IoU and mIoU.}
For class $c$, intersection over union is
\begin{equation}
\operatorname{IoU}_c=
\frac{\operatorname{TP}_c}
{\operatorname{TP}_c+\operatorname{FP}_c+\operatorname{FN}_c}.
\label{eq:metric_iou}
\end{equation}
For binary building extraction and optical change detection, we report the
foreground-class IoU. For multi-class flood and land-cover segmentation, mean
IoU averages class-wise IoU over the $C$ evaluated classes,
\[
\operatorname{mIoU}
=
\frac{1}{C}\sum_{c=1}^{C}\operatorname{IoU}_c.
\]

\paragraph{Height metrics.}
Let $\Omega$ be the set of valid evaluation pixels, and let $h_p$ and
$\hat h_p$ denote the reference and predicted height in meters. Potsdam is
evaluated with root-mean-square error,
\begin{equation}
\operatorname{RMSE}
=
\sqrt{
\frac{1}{|\Omega|}
\sum_{p\in\Omega}(\hat h_p-h_p)^2
},
\label{eq:metric_rmse}
\end{equation}
where lower is better. DFC23 Track~2 uses the threshold accuracy $\delta_1$,
i.e., the percentage of valid pixels whose multiplicative error is below $1.25$:
\begin{equation}
\delta_1=
\frac{100}{|\Omega|}
\sum_{p\in\Omega}
\mathbf{1}\!\left[
\max\!\left(
\frac{\max(\hat h_p,\epsilon)}{\max(h_p,\epsilon)},
\frac{\max(h_p,\epsilon)}{\max(\hat h_p,\epsilon)}
\right)<1.25
\right].
\label{eq:metric_delta}
\end{equation}
We use the official validity mask and set $\epsilon$ to avoid division by zero.

\paragraph{Referring-segmentation metrics.}
For query $i$, let $I_i$ and $U_i$ be the pixel intersection and union between
its predicted and reference masks. gIoU averages the per-query IoUs, whereas
cIoU pools pixels across the complete evaluation set:
\begin{equation}
\operatorname{gIoU}
=
\frac{100}{N}
\sum_{i=1}^{N}
\frac{I_i}{U_i},
\qquad
\operatorname{cIoU}
=
100
\frac{\sum_{i=1}^{N}I_i}
{\sum_{i=1}^{N}U_i}.
\label{eq:metric_ref_iou}
\end{equation}
gIoU weights every query equally, while cIoU gives greater influence to masks
with larger union areas. We use the same cIoU/gIoU names for both referring
benchmarks; these correspond to the oIoU/mIoU terminology, respectively, used
in the original RRSIS-D evaluation. For zero-shot transfer analyses, we
additionally report $\mathrm{Pr}@\alpha$, the percentage of queries whose
individual IoU is at least threshold $\alpha$.

\paragraph{Rotated-detection metric.}
For each DIOR-R class, AP50 is the area under the precision--recall curve when
a prediction is counted as a true positive at a rotated-box IoU threshold of
$0.5$. mAP50 is the unweighted mean of AP50 over the 20 object classes;
duplicate detections are treated as false positives by the standard one-to-one
matching procedure.

\FloatBarrier
\section{Structured-Target Codecs and Deterministic Recovery}
\label{app:target_codec}
This section specifies the deterministic visual carriers used for each task and evaluates how faithfully they survive the frozen VAE interface before any flow-model learning is involved.

\suppressfloats[t]
\begin{figure}[!t]
\centering
\includegraphics[width=\linewidth]{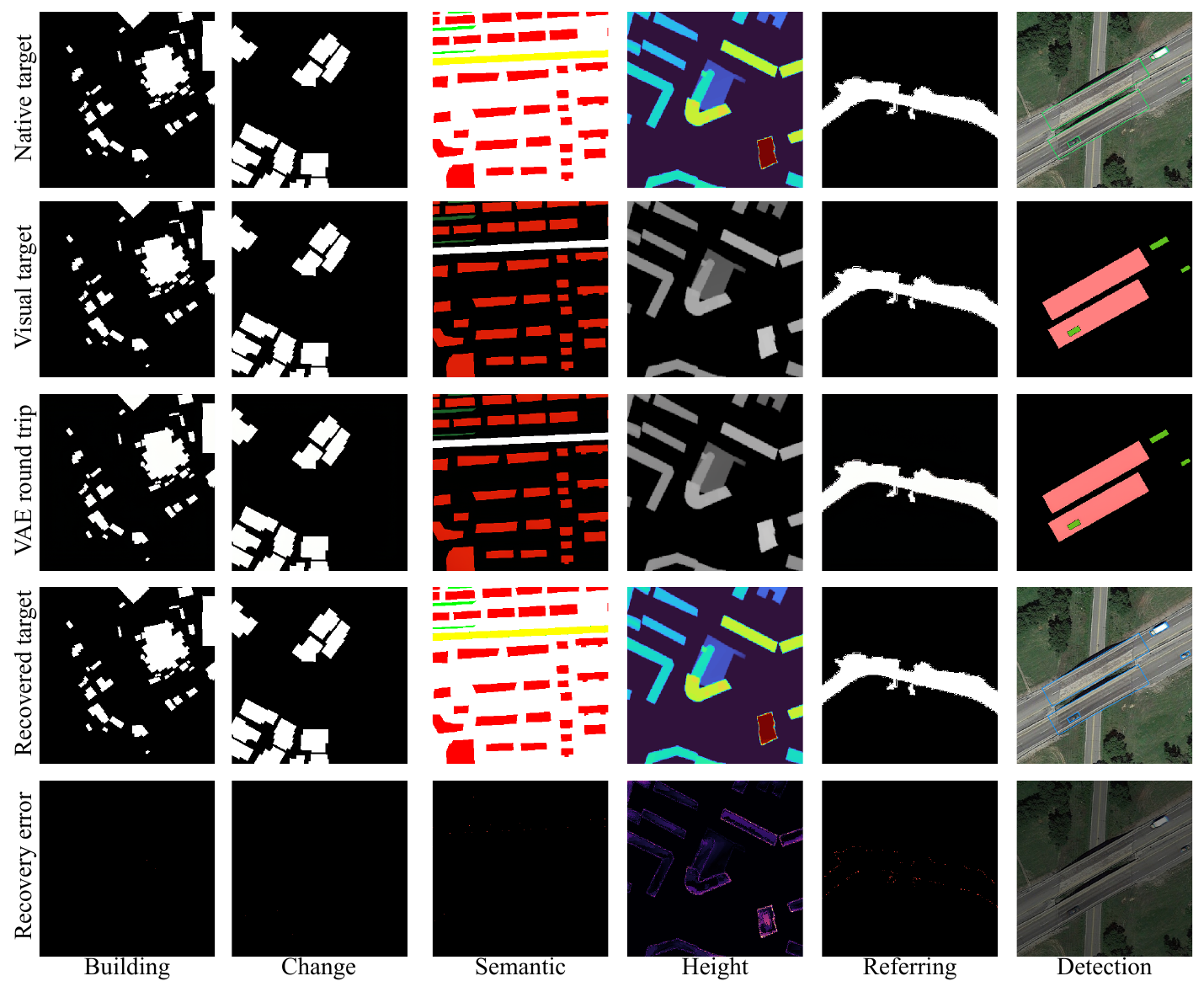}
\caption{
Frozen-VAE round-trip analysis of the structured visual target space.
Columns show building, change, semantic, height, referring, and rotated detection targets; rows show the native target, visual target, VAE round trip, recovered target, and recovery error.
The experiment measures whether task semantics are preserved by the frozen VAE independently of flow-model learning.
}
\label{fig:roundtrip}
\end{figure}

\paragraph{Binary targets.}
Building extraction, binary change detection, and referring segmentation use a black background and white foreground. This maximizes separation in the visual carrier while keeping recovery a deterministic thresholding operation.

\paragraph{Semantic targets.}
LoveDA and OpenEarthMap use a shared canonical semantic codebook. Dataset-native
class IDs are deterministically mapped into canonical semantics before rendering,
and evaluation maps the recovered predictions back to each dataset's native class
IDs. To determine the visual carrier, we compare three palette choices:
\emph{dataset official palettes}, which preserve each dataset's released
visualization colors; a \emph{shared canonical palette}, which uses one common
palette after ontology alignment; and our \emph{VAE-aware canonical palette},
which keeps the same canonical semantic mapping but selects the visual color codes
according to their frozen-VAE round-trip stability. UrbanSARFloods uses a separate
three-class flood encoding because its ontology is not a land-cover codebook.

\paragraph{Metric height targets.}
Potsdam and DFC23 share a log-compressed height representation over $[0,200]$ meters:
\begin{equation}
 u=\frac{\log(1+\operatorname{clip}(h,0,200))}{\log(201)},
 \qquad
 C_h(h)=(2u-1,2u-1,2u-1),
\label{eq:heightencode}
\end{equation}
with inverse mapping
\begin{equation}
 \hat h=\exp(\hat u\log(201))-1,
\end{equation}
where $\hat u=\operatorname{clip}((\hat c+1)/2,0,1)$ for decoded grayscale $\hat c\in[-1,1]$.
The three-channel image is only a VAE carrier; evaluation is performed on a continuous single-channel height map in meters.

\paragraph{Multi-class rotated detection targets.}
DIOR-R is rendered as a 20-class palette canvas. After VAE decoding, each pixel is assigned to its nearest palette code. We then separate connected components by class, extract minimum-area rotated rectangles, and apply deterministic rotated-box post-processing. This recovery path contains no learned parameters and makes representation error measurable at both raster and instance levels.

\paragraph{Direct codec sanity check.}
Before involving the frozen VAE, we verify the direct deterministic path
\begin{equation}
y_d^{\mathrm{codec}}=R_d(C_{\tau(d),d}(y_d)),
\qquad
B_d^{\mathrm{codec}}=M_d(y_d^{\mathrm{codec}},y_d).
\end{equation}
For binary masks, discrete semantic palettes, and valid-range height encodings, this check verifies that the evaluation implementation itself is consistent with the encoding definition. For DIOR-R, we report raster mIoU, mAP50, and split rate after connected components, minimum-area rectangle fitting, and rotated-box post-processing. These numbers quantify the geometry loss introduced by deterministic instance recovery independently of both the VAE and the flow model.

\paragraph{Frozen-VAE round-trip protocol.}
To isolate representation loss from flow-model prediction error, we evaluate
\begin{equation}
\tilde y_d=R_d\!\left(D\!\left(E\!\left(C_{\tau(d),d}(y_d)\right)\right)\right),
\qquad
B_d^{\mathrm{repr}}=M_d(\tilde y_d,y_d).
\label{eq:roundtrip}
\end{equation}
All codec comparisons follow the data-split protocol described in Appendix~\ref{app:dataset_details}.
The input resolution, VAE precision, and recovery operator are fixed, and only the visual encoding is varied.
This protocol isolates representation loss introduced by the frozen VAE and deterministic recovery.

\paragraph{Round-trip results.}
Figure~\ref{fig:roundtrip} visualizes the frozen-VAE round trip for all six
structured output families.

Binary masks remain visually stable through the round trip, while the error maps
make the more subtle palette and geometric deviations in semantic, height, and
rotated-instance targets explicit. We next quantify the semantic, height, and
rotated-instance visual carriers under the same frozen-VAE round-trip protocol.

\begin{table}[H]
\caption{Semantic-palette ablation under frozen-VAE round-trip evaluation. Results are reported as native mIoU.}
\label{tab:palette}
\centering
\tablebodyfont
\begin{tabular}{lcccc}
\toprule
Semantic target palette
& LoveDA $\uparrow$
& OEM-RGB $\uparrow$
& OEM-SAR $\uparrow$
& Mean $\uparrow$ \\
\midrule
Dataset official palettes
& 99.41 & 96.57 & 90.79 & 95.59 \\
Shared canonical palette
& \textbf{99.73} & 93.28 & 87.91 & 93.64 \\
\textbf{VAE-aware canonical (ours)}
& 99.72 & \textbf{98.83} & \textbf{99.11} & \textbf{99.22} \\
\bottomrule
\end{tabular}
\end{table}

\paragraph{Semantic-palette ablation.}
Table~\ref{tab:palette} shows that simply sharing a canonical palette does not
guarantee good compatibility with the frozen VAE: its mean round-trip mIoU is
93.64. Selecting the canonical visual codes according to frozen-VAE round-trip
stability raises the mean to 99.22, while retaining a shared semantic codebook
across LoveDA and OpenEarthMap. This supports the use of the VAE-aware canonical
palette as the semantic visual carrier in the main model.

\begin{table}[H]
\caption{Height-encoding ablation under frozen-VAE round-trip evaluation. Both variants use the same frozen VAE and fixed 0.2-m ground-recovery threshold.}
\label{tab:heightcodec}
\centering
\tablebodyfont
\begin{tabular}{lcc}
\toprule
Height encoding & Potsdam RMSE (m) $\downarrow$ & DFC23 $\delta_1$ (\%) $\uparrow$ \\
\midrule
Linear grayscale & 0.378 & 72.22 \\
\textbf{Log1p grayscale (ours)} & \textbf{0.176} & \textbf{97.42} \\
\bottomrule
\end{tabular}
\end{table}

Table~\ref{tab:heightcodec} compares linear and log-compressed
height encodings under the frozen-VAE round-trip protocol.
Log-compressed encoding reduces Potsdam RMSE from $0.378$
to $0.176$ m and improves DFC23 $\delta_1$ from $72.22\%$
to $97.42\%$, demonstrating better preservation of
continuous height information through the frozen VAE.

\begin{table}[H]

\caption{
\rev{DIOR-R raster and instance recovery fidelity.
Direct recovery is compared with frozen-VAE
round-trip recovery using ground-truth visual targets.}
}

\label{tab:diorcodec}
\centering
\tablebodyfont
\setlength{\tabcolsep}{5pt}
\begin{tabular}{llccc}
\toprule
Palette & Recovery path & Raster mIoU $\uparrow$ & mAP50 $\uparrow$ & Split $\downarrow$ \\
\midrule
VAE-aware hybrid & Direct recovery & \textbf{100.00} & 89.13 & \textbf{0.14} \\
\midrule
Initial palette & VAE + recovery & 97.90 & 89.44 & 0.50 \\
\textbf{VAE-aware hybrid palette (ours)}
& VAE + recovery & 98.43 & \textbf{90.10} & 0.44 \\
\bottomrule
\end{tabular}
\end{table}

Table~\ref{tab:diorcodec} examines the effects of
deterministic instance recovery and frozen-VAE
reconstruction on DIOR-R.
\rev{Direct recovery achieves perfect raster fidelity
(100.00 mIoU) but yields 89.13 mAP50, revealing
geometric losses during instance recovery.
Among the VAE round-trip variants, the VAE-aware
hybrid palette improves raster mIoU from 97.90
to 98.43 and mAP50 from 89.44 to 90.10,
while reducing the split rate from 0.50 to 0.44.
Compared with direct recovery, the VAE round trip
reduces raster fidelity and increases instance
splitting, despite slightly higher aggregate mAP50.
Figure~\ref{fig:recovery_fidelity} further examines
these effects on small instances.}

\subsection{Small-Instance Geometry and Semantic Boundary Fidelity}
\label{app:recovery_fidelity}

\rev{Following the aggregate recovery analysis in
Table~\ref{tab:diorcodec}, we further examine
geometric fidelity for small instances and fine
semantic boundaries.
We compare direct recovery, $R_d(C_d(y_d))$,
with VAE round-trip recovery,
$R_d(D(E(C_d(y_d))))$, where $C_d$ abbreviates
$C_{\tau(d),d}$.
Both paths start from ground-truth targets and use
the same deterministic recovery procedure.}

\begin{figure}[!htbp]
\centering
\includegraphics[width=\linewidth]{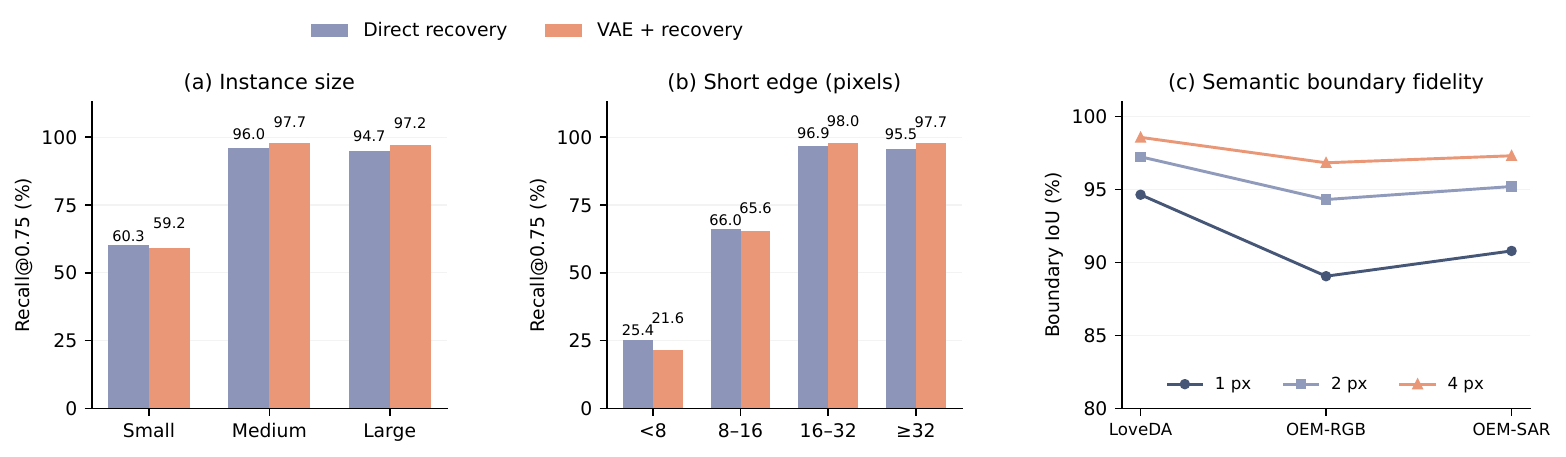}
\caption{
\rev{Recovery fidelity for small instances and semantic boundaries.
(a,b) DIOR-R recall at polygon IoU $0.75$, grouped by
instance area and rotated-box short-edge length,
comparing direct and VAE round-trip recovery.
(c) Class-macro boundary IoU after VAE round-trip
recovery at 1-, 2-, and 4-pixel widths.
Direct recovery achieves $100\%$ boundary IoU
and is omitted from (c).}
}
\label{fig:recovery_fidelity}
\end{figure}

\rev{\paragraph{Small-instance geometry.}
Figure~\ref{fig:recovery_fidelity}(a,b) shows
that recall at IoU $0.75$ decreases from
$60.32\%$ to $59.21\%$ for small instances
(area $<32^2$ pixels), and from $25.39\%$
to $21.56\%$ for instances with short edges
below 8 pixels after the VAE round trip.
For instances matched by both recovery paths
at IoU $0.5$, mean angular error in the latter
group increases from $1.017^\circ$ to
$1.148^\circ$, measured modulo $180^\circ$
and excluding aspect ratios below $1.1$.
These results show that thin instances are
particularly sensitive to VAE reconstruction,
with larger losses in geometric recall.}

\rev{\paragraph{Semantic boundary fidelity.}
We evaluate 1,669 LoveDA, 500 OEM-RGB, and
490 OEM-SAR images using the VAE-aware
canonical palette.
Boundary IoU is computed from class-wise inner
bands of 1, 2, and 4 pixels obtained through
repeated $3\times3$ erosion, with intersections
and unions aggregated across images before
class averaging.
Void neighborhoods and image borders are excluded.}

\rev{As shown in Figure~\ref{fig:recovery_fidelity}(c),
one-pixel boundary IoU reaches $94.64\%$,
$89.06\%$, and $90.79\%$ on LoveDA, OEM-RGB,
and OEM-SAR, respectively, increasing to
$98.57\%$, $96.83\%$, and $97.31\%$ at
four pixels.
These results indicate that fine boundaries
are more sensitive to VAE reconstruction
than broader semantic regions.
Together, the analyses reveal the remaining
geometric limitations of the shared visual
representation despite its high aggregate fidelity.}

\Needspace{7\baselineskip}
\section{Additional Analyses}
This section complements the main-text analysis with detailed source-state results,
transport-state visualizations, cross-task interaction, and zero-shot transfer studies.

\subsection{Source-State Initialization Details}
\label{sec:source_ablation_details}

Figure~\ref{fig:source_state} summarizes the source-state ablation in the main
text. Table~\ref{tab:source_full} reports the corresponding numerical results.
Visual conditioning is fixed within each panel so that only the source-state
initialization changes.

\begin{table}[H]
\caption{Full numerical results for the source-state initialization ablation.
Visual conditioning is fixed within each panel.}
\label{tab:source_full}
\centering
\tablebodyfont
\setlength{\tabcolsep}{5pt}
\textit{(a) Single-image tasks}\par\vspace{1pt}
\begin{tabular}{lccccc}
\toprule
Source $z_s$
& Bldg. IoU $\uparrow$
& Sem. mIoU $\uparrow$
& Height $\delta_1 \uparrow$
& Ref. gIoU $\uparrow$
& Det. mAP50 $\uparrow$ \\
\midrule
$\epsilon\sim\mathcal{N}(0,I)$
& 74.17 & 49.93 & 69.87 & 71.22 & 61.65 \\
\textbf{Current image latent $z_a$}
& \textbf{76.93} & \textbf{52.57} & \textbf{73.31}
& \textbf{73.93} & \textbf{64.75} \\
\bottomrule
\end{tabular}

\vspace{4pt}
\textit{(b) Bi-temporal change detection}\par\vspace{1pt}
\begin{tabular}{lcccc}
\toprule
Source $z_s$
& WHU-CD IoU $\uparrow$
& LEVIR-CD+ IoU $\uparrow$
& UrbanSAR mIoU $\uparrow$
& Mean $\uparrow$ \\
\midrule
$\epsilon\sim\mathcal{N}(0,I)$
& 86.49 & 72.26 & 52.32 & 70.36 \\
$z_a$
& 88.91 & 74.49 & 54.23 & 72.54 \\
$|z_a-z_b|$
& 90.14 & 75.70 & 55.61 & 73.82 \\
\textbf{$z_a-z_b$}
& \textbf{90.88} & \textbf{76.44} & \textbf{56.57} & \textbf{74.63} \\
\bottomrule
\end{tabular}
\end{table}

The numerical results confirm the trends summarized in
Fig.~\ref{fig:source_state}. For single-image tasks, $z_a$ improves over
Gaussian initialization on all five task metrics. For change detection,
the mean score increases from 70.36 with Gaussian initialization to 72.54 with
$z_a$, 73.82 with $|z_a-z_b|$, and 74.63 with the signed residual $z_a-z_b$.

\subsection{Transport-State Trajectories}
\label{app:transport_trajectories}

\begin{figure}[H]
\centering
\includegraphics[width=0.87\linewidth]{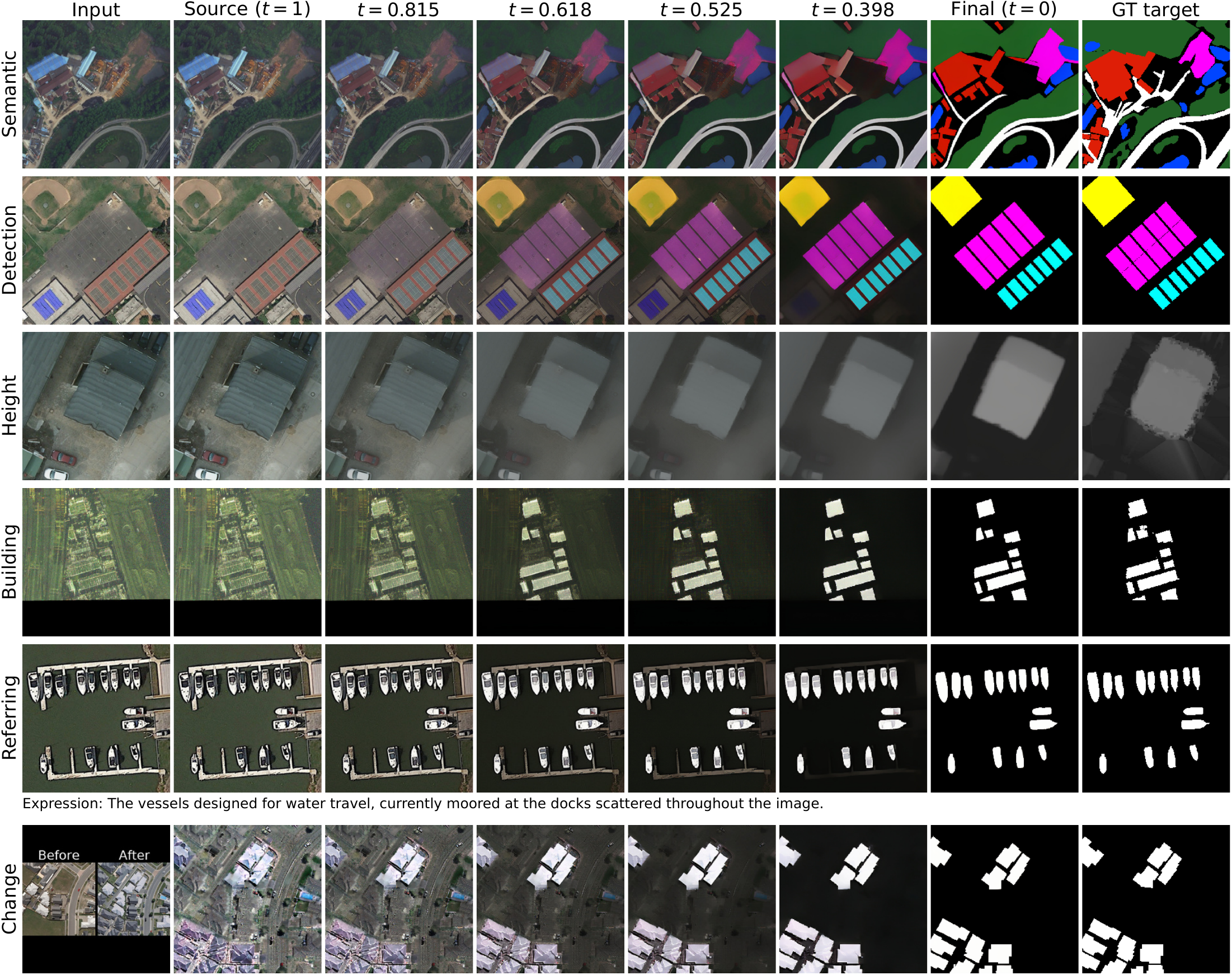}
\caption{\textbf{Transport-state trajectories across six tasks.}
All rows use the same shared \method{} checkpoint. From top to bottom:
semantic segmentation (LoveDA), rotated detection (DIOR-R), height estimation
(Potsdam), building extraction (SpaceNet~6 SAR), referring segmentation
(LaSeRS), and change detection (WHU-CD). Columns show the input, decoded
source state, selected intermediate states $D(z_t)$, final decoded state,
and ground-truth structured visual target.
}
\label{fig:transport_trajectories}
\end{figure}

Figure~\ref{fig:transport_trajectories} visualizes how the shared predictor
transports task-dependent source states toward structured visual outputs.
We use the same trained \method{} checkpoint for all six tasks and run ten-step
Euler inference to expose intermediate states along the trajectory. The main
quantitative evaluation continues to use the default two-step inference setting.
For each example, we show solver states at $t=1$, $0.815$, $0.618$, $0.525$,
$0.398$, and $0$, with times rounded for presentation.

\paragraph{Decoded transport states.}
Each intermediate panel directly decodes the current solver state as $D(z_t)$
using the frozen VAE decoder. No thresholding, palette snapping, or instance
recovery is applied to these intermediate outputs, and the ground-truth
structured target is shown only in the rightmost column for reference.
Accordingly, height estimation is visualized in its log-compressed grayscale
representation, while rotated detection is shown as the generated class-coded
canvas before geometric instance recovery. For change detection, the source
state is the signed latent difference $z_a-z_b$ (after minus before), so its
initial decoding differs from the appearance of either input observation.
Referring segmentation uses the sample's original expression, shown below the
corresponding row.

Across the single-image tasks, the decoded states progressively evolve from
image-like source representations toward binary, categorical, or continuous
target structures. Change detection follows a different trajectory from its
latent-difference source. Together, these examples visualize how the same
latent flow process produces structurally different outputs across tasks.

\FloatBarrier
\Needspace{7\baselineskip}

\subsection{Cross-Task Interaction}
\label{sec:interaction}

We analyze task gradients on the shared LoRA to understand how different tasks interact during joint training. At the same initialization checkpoint, we compute the gradient of each task loss with respect to the shared LoRA parameters $\phi$ and measure the cosine similarity between task pairs:
\begin{equation}
g_\tau = \nabla_{\phi}\mathcal{L}_\tau,
\qquad
\rho_{\tau,j}
=
\frac{g_\tau^\top g_j}
{\lVert g_\tau\rVert_2\,\lVert g_j\rVert_2}.
\label{eq:interaction}
\end{equation}
Positive similarity indicates that two tasks favor aligned parameter updates, while negative similarity indicates conflicting update directions.

We compare \method{} with a task-specific-head variant. Both use the same frozen FLUX.2/MMDiT backbone, shared LoRA configuration, conditioning inputs, initialization checkpoint, and TMAS setting. The difference lies in their prediction interfaces: \method{} uses a unified generative predictor, whereas the variant employs independent task-specific heads.

\begin{figure}[H]
\centering
\includegraphics[width=0.8\linewidth]{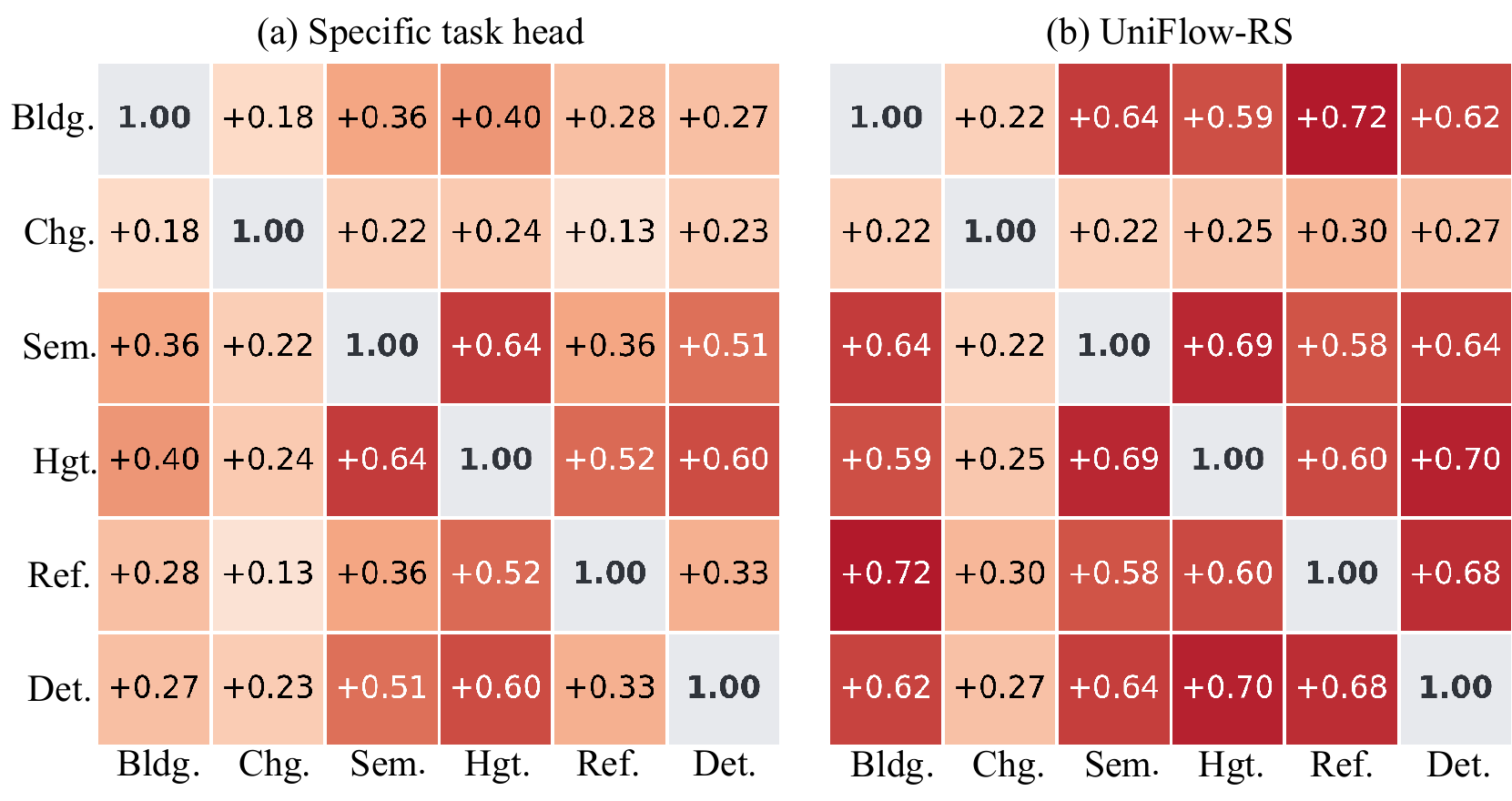}
\caption{
Pairwise gradient cosine similarities between tasks on the shared LoRA at the same initialization checkpoint: (a) task-specific heads and (b) \method{}. Diagonal entries are 1 by construction.
}
\label{fig:gradient_interaction}
\end{figure}

Figure~\ref{fig:gradient_interaction} shows positive gradient similarities for all fifteen task pairs. \method{} achieves a higher mean off-diagonal similarity than the task-specific-head variant ($0.51$ vs. $0.35$), indicating greater alignment among task gradients. Building--referring ($0.72$) and height--detection ($0.70$) exhibit the strongest alignment, whereas change detection shows weaker alignment with the other tasks ($0.22$--$0.30$). These findings suggest that the unified prediction formulation encourages more compatible task updates at the measured checkpoint, consistent with the joint-training improvements in Table~\ref{tab:joint}.

\Needspace{7\baselineskip}
\subsection{Zero-Shot Transfer Details}
\label{sec:transfer_details}
We extend the main-text zero-shot evaluation with thresholded overlap precision, which measures how often individual queries remain accurate under progressively stricter IoU criteria.

\begin{table}[H]
\caption{Detailed zero-shot transfer results on two unseen datasets. $\mathrm{Pr}@\alpha$ denotes the percentage of queries with IoU $\geq \alpha$, and ``--'' indicates that the corresponding result is not reported.}
\label{tab:generalization_full}
\centering
\tablebodyfont
\setlength{\tabcolsep}{3.7pt}
\begin{tabular}{llccccccc}
\toprule
Target & Method & cIoU & gIoU & Pr@0.5 & Pr@0.6 & Pr@0.7 & Pr@0.8 & Pr@0.9 \\
\midrule
\multirow{3}{*}{EarthReason}
& FLUX.2-Klein Base & 17.74 & 17.04 & 8.92 & 4.15 & 1.97 & 0.78 & 0.05 \\
& RS2-SAM2-RRSISD & 35.48 & 36.55 & 37.14 & 32.47 & 28.01 & 22.51 & 14.63 \\
& \textbf{\method{}} & \textbf{44.23} & \textbf{48.24} & \textbf{50.88} & \textbf{45.64} & \textbf{41.65} & \textbf{35.79} & \textbf{24.74} \\
\midrule
\multirow{3}{*}{RISBench}
& FLUX.2-Klein Base & 6.38 & 7.74 & 2.78 & 1.44 & 0.75 & 0.29 & 0.07 \\
& VisionReasoner & 46.98 & 51.62 & -- & -- & -- & -- & -- \\
& \textbf{\method{}} & \textbf{70.92} & \textbf{64.62} & \textbf{72.07} & \textbf{67.31} & \textbf{60.96} & \textbf{51.03} & \textbf{32.47} \\
\bottomrule
\end{tabular}
\end{table}

Table~\ref{tab:generalization_full} reports thresholded overlap precision in addition to cIoU and gIoU. On EarthReason, where both imagery and text queries are out of distribution with respect to joint training, the advantage of \method{} persists as the overlap criterion becomes stricter: $\mathrm{Pr}@0.9$ increases from 14.63 for RS2-SAM2-RRSISD to 24.74. RISBench, which primarily introduces a visual distribution shift, similarly retains 32.47 $\mathrm{Pr}@0.9$, complementing the unseen-dataset results reported in the main text.

\Needspace{7\baselineskip}

\subsection{Transfer under Canopy and Cloud Occlusion}
\label{sec:rsc_protocol}

\rev{We evaluate zero-shot transfer on Chesapeake RSC and PIE
using the same jointly trained \method{} checkpoint,
without target-specific adaptation.}

\rev{\paragraph{Chesapeake RSC protocol.}
We evaluate all 3,000 Chesapeake RSC test images
~\citep{robinson2024chesapeake} at $512\times512$ resolution,
including 466 images without annotated roads.
Only RGB inputs are used.
The ground truth includes both visible and canopy-occluded
roads, with unclassified and no-data pixels excluded.
IoU and F1 are computed from dataset-level aggregated
pixel counts.
We additionally report canopy-road recall,
$R_{\rm canopy}=\mathrm{TP}_{\rm canopy}/N_{\rm canopy}$,
to measure the recovery of roads beneath tree canopies.}

\rev{\paragraph{PIE evaluation protocol.}
We use the PIE-RGB-SAR dataset~\citep{zhang2024asanet}
from the processed release of MM-OVSeg~\citep{wei2026mmovseg},
evaluating all 2,432 held-out validation samples.
Evaluation covers four semantic classes:
road, water, forest, and farmland.
Pixels outside these classes remain negatives in the
corresponding class-wise evaluations.
For RGB-only models, the original $256\times256$ images
are resized to $512\times512$ after cloud corruption,
and predictions are restored to the native resolution
using nearest-neighbor interpolation.
Class-wise IoUs are computed from aggregated pixel counts.}

\rev{\paragraph{Synthetic cloud generation.}
We generate cloud corruption using SatelliteCloudGenerator,
which produces a Perlin-based opacity field $a(x)$.
The corrupted image is defined as
\[
I_s(x)=(1-sa(x))I(x)+sa(x),
\]
where $s\in\{0,0.45,0.90\}$ represents clear,
thin-cloud, and thick-cloud conditions, respectively.
The same opacity field is used across cloud levels
for each image.
Images are clipped to $[0,1]$ and converted to
8-bit RGB, preserving originally black pixels.
All methods receive the same corrupted RGB observations
at each cloud level; MM-OVSeg additionally receives
the paired uncorrupted SAR imagery.}

\rev{\paragraph{Baseline configurations.}
All models are evaluated with frozen checkpoints and
without target-domain adaptation.
S5 ViT-L~\citep{lv2026s5} uses its released LoveDA
segmentation head without test-time augmentation.
FLUX.2-Klein Base follows the same structured-target
inference protocol as \method{}, including text
conditioning, source-state initialization, frozen VAE,
two-step inference, BF16 precision, and deterministic
recovery, but without the trained LoRA adapter.}

\rev{For Chesapeake RSC, we use the automatic road-extraction
baseline released with RoadGIE~\citep{peng2026roadgie},
consisting of an SMT-base backbone and DaFormer-ASPP
decoder~\citep{hoyer2022daformer} with WorldRoadSeg-360K
weights.
The model receives no interactive prompts or target
annotations.
We retain its native $512\times512$ inputs, RGB
normalization, bilinear upsampling, FP32 precision,
and a fixed sigmoid threshold of 0.5.
A fixed evaluation adapter is used to evaluate the
released checkpoint, with strict weight loading.}

\rev{For PIE, we evaluate MM-OVSeg B16~\citep{wei2026mmovseg}
using its released \texttt{modelDDSK.pth} checkpoint
trained on DDHR-SK.
We use the official six-class PIE vocabulary,
merging background and city predictions into Other,
which remains in the confusion matrix but is excluded
from the four-class mIoU.
Following its released preprocessing, RGB images are
resized to $384\times384$, while the native
$256\times256$ SAR images are padded to the same size
with value 128.
Predictions are restored to the native evaluation grid
before scoring.
The model runs in FP32 without PIE-specific adaptation.}

\Needspace{8\baselineskip}
\subsection{Transfer to Near-Infrared False-Color Imagery}
\label{sec:irrg_transfer}

\rev{We further evaluate zero-shot building extraction on
ISPRS Vaihingen~\citep{rottensteiner2012isprs},
whose false-color imagery contains near-infrared,
red, and green (IRRG) channels.
This setting examines the ability of the shared
generative predictor to transfer to a different
spectral composition without target-specific adaptation.}

\begin{table}[H]
\caption{
\rev{Zero-shot building extraction on Vaihingen IRRG imagery.
Results are aggregated over 17 held-out areas (398 patches).
All metrics are reported as percentages.}
}
\label{tab:vaihingen_irrg}
\centering
\tablebodyfont
\setlength{\tabcolsep}{7pt}
\begin{tabular}{@{}lcccc@{}}
\toprule
Method & IoU $\uparrow$ & F1 $\uparrow$ & Precision $\uparrow$ & Recall $\uparrow$ \\
\midrule
FLUX.2-Klein Base
& 19.56 & 32.72 & 40.99 & 27.22 \\
\methodsrc{li2025segearthov}{SegEarth-OV}{CVPR'25}
& 53.00 & 69.28 & 60.02 & 81.91 \\
\methodsrc{zhang2025corrclip}{CorrCLIP}{ICCV'25}
& 49.06 & 65.83 & 50.09 & \textbf{95.99} \\
\textbf{\method{}}
& \textbf{86.59} & \textbf{92.81}
& \textbf{95.57} & 90.21 \\
\bottomrule
\end{tabular}
\end{table}

\rev{\paragraph{Evaluation protocol.}
We evaluate all 17 held-out Vaihingen areas,
partitioned into 398 nonoverlapping $512\times512$ patches.
The original 8-bit IRRG channels are preserved
without band substitution or intensity stretching.
Padded image boundaries are excluded from evaluation.
All non-building classes are treated as background,
and patches without buildings are retained.
We evaluate binary building extraction without
boundary erosion.
All methods use the same input imagery and valid-pixel
masks, with model-specific preprocessing applied
as required.}

\rev{\paragraph{Baseline configurations.}
All models use frozen checkpoints without target-domain
adaptation.
\method{} and FLUX.2-Klein Base share the same fixed
building prompt, two-step inference, $512\times512$
resolution
and binary threshold of 0.5.}

\rev{SegEarth-OV~\citep{li2025segearthov} follows its released
Vaihingen configuration, using CLIP ViT-B/16,
SimFeatUp weights, and a six-class vocabulary.
CorrCLIP~\citep{zhang2025corrclip} uses its released
implementation with MetaCLIP ViT-B/16,
DINO ViT-B/8, and SAM2 Hiera-large.
We adopt its dynamic SAM2$_8$ variant with
eight points per side and the default region-generation
and majority-vote procedures.
For Vaihingen, we use the same class vocabulary
as SegEarth-OV and the released ADE settings,
including a 448-pixel internal resize,
336-pixel crops, and a stride of 112 pixels.
Both baselines restore predictions to the
$512\times512$ grid before extracting building masks.}

\rev{All prompts, thresholds, and model configurations
remain fixed during evaluation.
The available S5 checkpoint is excluded because
it includes Vaihingen downstream training.
Pretraining data and model capacities differ
across the compared methods.}

\rev{\paragraph{Results.}
As shown in Table~\ref{tab:vaihingen_irrg},
\method{} achieves 86.59 IoU and 92.81 F1,
outperforming SegEarth-OV and CorrCLIP by
33.59 and 37.53 IoU points, respectively.
It also achieves 95.57 precision while maintaining
90.21 recall, substantially reducing false positives
compared with the baselines.
These results suggest that the structural priors learned by UniFlow-RS are not confined to the RGB and SAR observations used during training. The shared generative predictor can effectively recover building structures from near-infrared false-color imagery, highlighting its potential for zero-shot transfer across spectral distributions.}

\section{TMAS Implementation and Component Ablations}
\label{app:tmas_details}

\subsection{Estimator and Adaptive Allocation}

We provide the implementation details of TMAS, including group-wise error estimation, distributed aggregation, handling of unobserved timestep bins, numerical stabilization, and adaptive sampling.

\paragraph{Group keys and per-sample error.}
TMAS maintains a sampling profile for each task--modality group $g=(\tau,m)$, where $\tau$ denotes one of the six tasks and $m$ denotes the physical modality of the training sample. Datasets sharing the same task and modality contribute to a common profile, while RGB and SAR observations are assigned to separate profiles. For multimodal datasets, the group is determined by the modality of the sampled view.

The resulting ten groups are
\[
\begin{aligned}
\mathcal{G}=\{&\texttt{building/rgb},\ \texttt{building/sar},\
                \texttt{change/rgb},\ \texttt{change/sar},\\
              &\texttt{semantic/rgb},\ \texttt{semantic/sar},\
                \texttt{height/rgb},\ \texttt{height/sar},\\
              &\texttt{referring/rgb},\ \texttt{detection/rgb}\}.
\end{aligned}
\]
For example, WHU-CD and LEVIR-CD+ jointly update the \texttt{change/rgb} profile.

For each sample $i$, the estimator computes the token-normalized velocity-prediction error:
\begin{equation}
\ell_i=
\frac{\sum_{l,c}w_{i,l}
\left(v_{\theta,i,l,c}-v^\star_{i,l,c}\right)^2}
{\max(D_z\sum_l w_{i,l},1)},
\label{eq:per_sample_loss}
\end{equation}
where $l$ and $c$ index packed latent tokens and feature dimensions, respectively, $w_{i,l}$ denotes the valid-token weight, and $D_z$ is the token feature dimensionality. When no validity mask is used, $w_{i,l}=1$. Each sample is assigned to timestep bin $k_i=\min(\lfloor Kt_i\rfloor,K-1)$.

\Needspace{11\baselineskip}
\paragraph{Distributed group--bin aggregation.}
Let $\mathcal B_r^\circ$ denote the samples with finite estimator losses on worker $r$. The group--bin mean error is computed by aggregating loss sums and sample counts across workers:
\begin{equation}
\bar\ell_{g,k}=
\frac{\sum_r\sum_{i\in\mathcal B_r^\circ}
\mathbf 1[g_i=g,\ k_i=k] \, \ell_i}
{\sum_r\sum_{i\in\mathcal B_r^\circ}
\mathbf 1[g_i=g,\ k_i=k]},
\label{eq:group_bin_mean}
\end{equation}
whenever at least one sample contributes to the group--bin pair. Each EMA state is initialized from its first valid observation and subsequently updated according to Eq.~\ref{eq:difficulty}. Updates occur only when the corresponding group--bin pair is observed in a microbatch.

\paragraph{Stabilized difficulty and activity.}
For unobserved timestep bins, we initialize the sampling-time estimates using the mean slow-EMA value of the observed bins within the same group. Specifically, if $\mathcal O_g$ is the set of observed bins, each unobserved bin receives $|\mathcal O_g|^{-1}\sum_{j\in\mathcal O_g}d^{\mathrm{slow}}_{g,j}$ in both EMA tables. Groups with no observations use a fallback value of $1$. All estimates are lower-bounded by $\epsilon$, yielding $\tilde d^{\mathrm{fast}}_{g,k}$ and $\tilde d^{\mathrm{slow}}_{g,k}$. Assigning the same fallback to both EMA tables gives zero learning activity for unobserved bins.

Using these stabilized estimates, the difficulty and learning-activity signals are computed as
\begin{equation}
\begin{aligned}
D_{g,k}
&=
\frac{\tilde d^{\mathrm{slow}}_{g,k}}
{K^{-1}\sum_j\tilde d^{\mathrm{slow}}_{g,j}},\\
A_{g,k}
&=
\min\!\left\{
\left|
\frac{\tilde d^{\mathrm{slow}}_{g,k}-\tilde d^{\mathrm{fast}}_{g,k}}
{\tilde d^{\mathrm{slow}}_{g,k}}
\right|,
1
\right\}.
\end{aligned}
\label{eq:app_learning_state}
\end{equation}
$D_{g,k}$ measures timestep difficulty relative to the group average, while $A_{g,k}$ captures recent learning activity through the absolute fast--slow EMA discrepancy, accounting for both improving and degrading regions.

For the improvement-only activity ablation, we define
\[
r_{g,k}=\frac{\tilde d^{\mathrm{slow}}_{g,k}-\tilde d^{\mathrm{fast}}_{g,k}}{\tilde d^{\mathrm{slow}}_{g,k}},
\]
and use $A_{g,k}=\min(\max(r_{g,k},0),1)$ instead of the default $A_{g,k}=\min(|r_{g,k}|,1)$.

\paragraph{Utility and numerical safeguards.}
The sampling utility is computed as
\begin{equation}
U_{g,k}
=
\max\!\left\{
(D_{g,k}+\epsilon)\left(\frac{1}{K}+A_{g,k}\right),
\,\epsilon
\right\}.
\label{eq:app_utility}
\end{equation}
The lower bound extends Eq.~\ref{eq:utility} with a numerical safeguard. Non-finite relative changes are replaced by zero, and non-finite utilities are replaced by one before normalization. For a group with no observations, all bins have equal utility, resulting in uniform sampling.

\paragraph{Bin and timestep sampling.}
Following Eq.~\ref{eq:tmas}, we first sample a timestep bin and then draw a continuous timestep uniformly within the selected bin:
\begin{equation}
k\sim\operatorname{Categorical}(p_s(\cdot\mid g)),
\qquad
t\mid k\sim\mathcal{U}\!\left[\frac{k}{K},\frac{k+1}{K}\right).
\label{eq:bin_sampling}
\end{equation}
Training progress is defined as $s=\operatorname{clip}(n/N,0,1)$, where $n$ is the number of completed optimizer updates and $N$ is the total training budget. Sampling remains uniform for $s\leq 1/K$, after which the group-specific utility mixture determines the timestep distribution. The distribution is updated using online EMA statistics and renormalized for numerical stability.

\paragraph{Default settings and reproducibility.}
TMAS uses two sampler hyperparameters: the number of timestep bins $K$ and the exploration weight $\lambda$. We set $K=10$, $\lambda=0.2$, and $\epsilon=10^{-6}$. The remaining coefficients are determined by $K$: the fast and slow EMA decay factors are $1-1/K=0.90$ and $1-1/K^2=0.99$, respectively; the utility offset is $1/K=0.1$; and the initial uniform phase spans $1/K=0.1$ of training (6k steps under the 60k-step schedule). The activity signal is bounded to $[0,1]$.

With $\lambda=0.2$, each timestep bin retains a minimum sampling probability of $\lambda/K=0.02$. The sampler state is detached from the optimization graph and synchronized across workers. Task and dataset sampling weights remain fixed across all sampler ablations.

\Needspace{23\baselineskip}

\subsection{Ablation of Learning Activity and Exploration}
\label{app:tmas_components}

We examine the contributions of learning-activity estimation and uniform exploration in \tmas{}. The former captures both improving and degrading timestep regions through the absolute fast--slow EMA discrepancy, while the latter provides initial coverage and continued exploration throughout training. We ablate these components to evaluate their effects on joint optimization.

\begin{table}[H]
\caption{
Ablation of learning activity and exploration in \tmas{}.
Improvement-only activity considers only decreasing errors.
No uniform warm-up removes the initial uniform sampling phase,
while no persistent exploration removes the uniform mixture after warm-up.
Steps denotes the updates required to reach the final UTS of Fixed Beta.
}
\label{tab:learning_state}
\centering
\tablebodyfont
\setlength{\tabcolsep}{4pt}
\begin{tabular}{lcccccccc}
\toprule
Variant
& Bldg. $\uparrow$
& Change $\uparrow$
& Sem. $\uparrow$
& Height $\uparrow$
& Ref. $\uparrow$
& Det. $\uparrow$
& UTS $\uparrow$
& Steps $\downarrow$ \\
\midrule

Improvement-only activity
& 76.79 & 74.47 & 52.24 & 73.06 & 73.77 & 64.50 & 69.14 & 42k \\

No uniform warm-up
& 76.58 & 74.29 & 51.96 & 72.84 & 73.51 & 64.22 & 68.90 & 49k \\

No persistent exploration ($\lambda=0$)
& 76.71 & 74.38 & 52.08 & 72.96 & 73.64 & 64.36 & 69.02 & 45k \\

\textbf{\tmas{}}
& \textbf{76.93}
& \textbf{74.63}
& \textbf{52.57}
& \textbf{73.31}
& \textbf{73.93}
& \textbf{64.75}
& \textbf{69.35}
& \textbf{37k} \\

\bottomrule
\end{tabular}
\end{table}

As shown in Table~\ref{tab:learning_state}, all three components contribute to the performance of \tmas{}. Restricting learning activity to improving regions reduces UTS from 69.35 to 69.14, suggesting that increasing-error regions also benefit from additional sampling. Removing uniform warm-up or persistent exploration further reduces UTS to 68.90 and 69.02, respectively. These variants also require more training steps to reach the Fixed Beta reference performance (42k--49k versus 37k). Together, the results highlight the complementary benefits of tracking bidirectional learning-state changes and maintaining timestep exploration during joint training.

\FloatBarrier
\Needspace{21\baselineskip}

\subsection{Task--Modality Timestep Allocation and Validation Dynamics}
\label{app:tmas-allocation-dynamics}

\begin{figure}[h]
    \centering
    \includegraphics[width=\linewidth]{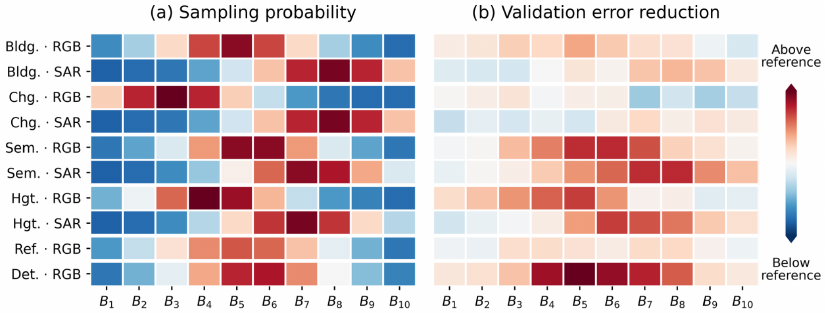}
    \caption{\rev{
    Task--modality-specific timestep allocation and validation-error changes.
    (a) TMAS sampling probabilities relative to uniform sampling ($1/K$).
    (b) Per-bin reduction in validation Flow Matching error relative to Fixed Beta.
    Warmer colors indicate higher relative sampling probabilities in (a)
    and larger error reductions in (b).}
    }
    \label{fig:tmas-allocation-dynamics}
\end{figure}

\rev{We further analyze how \tmas{} allocates timesteps across
heterogeneous task--modality groups and how these allocations
relate to validation performance.
Figure~\ref{fig:tmas-allocation-dynamics} compares the adaptive
sampling distributions with changes in validation Flow Matching
error relative to Fixed Beta across ten task--modality groups.}

\rev{Figure~\ref{fig:tmas-allocation-dynamics} (a) reveals distinct
timestep allocation profiles across tasks and between RGB and SAR
modalities.
The corresponding validation-error changes in
Figure~\ref{fig:tmas-allocation-dynamics} (b) are also heterogeneous.
Semantic segmentation and rotated detection exhibit larger error
reductions in selected timestep regions, whereas change detection
and referring segmentation show smaller or mixed changes.
These patterns are consistent with the task-dependent performance
gains in Table~\ref{tab:sampler}, supporting the use of
group-specific timestep allocation for heterogeneous joint training.}

\section{Capacity and Efficiency}
\label{app:efficiency}
The main text focuses on predictive sharing and optimization. This section reports the practical cost of that sharing separately, including adapter capacity and wall-clock behavior.

\subsection{Adapter Capacity}

\begin{table}[H]
\caption{LoRA capacity and efficiency. Rank 32 is the default used in all other experiments and is highlighted in the Rank column. Bold metric values indicate the best result in each column.}
\label{tab:lora}
\centering
\tablebodyfont
\renewcommand{\arraystretch}{0.96}
\begin{tabular}{ccccc}
\toprule
Rank & Trainable params $\downarrow$ & UTS $\uparrow$ & Peak memory $\downarrow$ & Images/s $\uparrow$ \\
\midrule
8 & \textbf{12.3M} & 67.24 & \textbf{40.92 GiB} & \textbf{6.69} \\
16 & 24.6M & 68.91 & 41.21 GiB & \textbf{6.69} \\
\textbf{32} & 49.1M & 69.35 & 41.53 GiB & \textbf{6.69} \\
64 & 98.3M & \textbf{69.37} & 42.23 GiB & \textbf{6.69} \\
\bottomrule
\end{tabular}
\vspace{2pt}

\end{table}

Table~\ref{tab:lora} reports the capacity--efficiency trade-off for the shared LoRA adapter. Increasing the rank from 8 to 32 raises UTS from $67.24$ to $69.35$, while doubling the rank from 32 to 64 yields only a $0.02$-point gain ($69.37$). Trainable parameters grow from $12.3$M to $98.3$M and peak memory increases from $40.92$ to $42.23$ GiB, whereas the measured throughput remains unchanged at $6.69$ images/s. Rank 32 is therefore used as the default operating point in the other experiments.

\Needspace{9\baselineskip}
\subsection{Inference Efficiency on Referring Segmentation}
\label{sec:large_model_efficiency}

\rev{We evaluate the inference efficiency of \method{}
against two language-conditioned segmentation models,
GeoPixel-7B-RES~\citep{shabbir2025geopixel} and
LISA-7B-v1~\citep{lai2024lisa}, with
RS2-SAM2-RRSISD~\citep{rong2026rs2sam2}
as a lightweight task-specific reference.
Parameter counts include all loaded vision,
language, and segmentation components.}

\paragraph{Measurement protocol.}
\rev{All methods are evaluated on the same 200 unique
RRSIS-D image--expression pairs, sampled with seed
12345 independently of prediction quality.
Experiments are conducted sequentially on a single
GPU (NVIDIA RTX 6000 Ada),
using batch size one and four CPU threads.
Each model undergoes 30 warm-up requests followed
by three measured rounds, totaling 600 requests.}

\rev{CUDA-synchronized latency is measured from decoded
CPU $512\times512$ images and raw expressions to
the final CPU masks, including preprocessing,
data transfers, online text encoding, inference,
and output recovery.
Disk I/O, weight loading, scoring, and visualization
are excluded.
Peak memory denotes PyTorch allocated memory,
including resident model weights.
Throughput is measured using serial batch-one inference.}

\begin{table}[H]
\caption{
\rev{End-to-end referring-segmentation efficiency on RRSIS-D.
Parameters include all loaded components.
Latency is reported in milliseconds and peak memory in GiB.
All methods process expressions online with batch size one.}
}
\label{tab:large_model_efficiency}
\centering
\tablebodyfont
\setlength{\tabcolsep}{6pt}
\begin{tabular}{@{}lccccc@{}}
\toprule
Method
& Params (B) $\downarrow$
& Mean $\downarrow$
& P95 $\downarrow$
& Memory $\downarrow$
& Samples/s $\uparrow$ \\
\midrule
RS2-SAM2-RRSISD
& \textbf{0.935}
& \textbf{234.90}
& \textbf{238.07}
& \textbf{4.018}
& \textbf{4.257} \\
LISA-7B-v1
& 7.705
& 638.48
& 721.41
& 15.992
& 1.566 \\
GeoPixel-7B-RES
& 11.338
& 6352.46
& 7691.27
& 27.193
& 0.157 \\
\method{}
& 8.032
& 321.22
& 322.35
& 13.291
& 3.113 \\
\bottomrule
\end{tabular}
\end{table}

\paragraph{Model configurations.}
\rev{\method{} uses two-step flow inference with a BF16
backbone and VAE, together with its resident FP8
text encoder without expression caching.
RS2-SAM2 uses FP32 precision and its native
SAM/BEiT resolutions of 1024/224 pixels.}

\rev{GeoPixel retains its released adaptive image
partitioning (\texttt{hd\_num=9}), SAM2,
BF16 precision, FlashAttention~2, and greedy
generation with a 300-token limit.
LISA uses BF16 precision, its native conversation
template, SAM/CLIP preprocessing at 1024/224 pixels,
and a 512-token generation limit.
Neither model receives ground-truth prompts or
answer tokens.
Across all measured requests, neither reaches
its generation limit or produces an empty mask.
Their mean generated lengths are 8.99 and 8.77
tokens, respectively.}

\rev{\paragraph{Results.}
As shown in Table~\ref{tab:large_model_efficiency},
\method{} achieves a mean latency of 321.22 ms
and a throughput of 3.113 samples/s,
making it 1.99$\times$ faster than LISA
and 19.78$\times$ faster than GeoPixel.
It also requires less peak memory than both models.
The lightweight RS2-SAM2 remains faster and
more memory-efficient, with 234.90 ms latency
and 4.018 GiB peak memory.}

\rev{These results show that a shared generative predictor
can achieve competitive end-to-end inference efficiency
compared with language-conditioned segmentation models,
while supporting multiple perception tasks through
a single checkpoint.}

\clearpage
\section{Full Benchmark Comparisons}
\label{app:full_benchmarks}
The main text uses Figure~\ref{fig:main_results} as a compact visual summary of performance across the thirteen evaluation sources. For completeness, Tables~\ref{tab:main_a} and~\ref{tab:main_b} report detailed numerical comparisons, including venue--year information and dataset-specific notes.

\begin{table}[H]
\caption{
Benchmark comparison for change detection, semantic segmentation, and rotated detection. Gray cells denote results reported by prior work, and bold indicates the best result in each column.
}
\label{tab:main_a}
\centering
\tablebodyfont
\setlength{\tabcolsep}{3.0pt}
\renewcommand{\arraystretch}{1.00}
\begin{tabular}{lccccccc}
\toprule
& \multicolumn{3}{>{\columncolor{taskChange}}c}{Change detection}
& \multicolumn{3}{>{\columncolor{taskSemantic}}c}{Semantic segmentation}
& \multicolumn{1}{>{\columncolor{taskDetection}}c}{Rotated detection} \\
\cmidrule(lr){2-4}\cmidrule(lr){5-7}\cmidrule(lr){8-8}
Method
& \shortstack{WHU-CD\\IoU $\uparrow$}
& \shortstack{LEVIR-CD+\\IoU $\uparrow$}
& \shortstack{UrbanSAR\\mIoU $\uparrow$}
& \shortstack{LoveDA\\mIoU $\uparrow$}
& \shortstack{OEM-RGB\\mIoU $\uparrow$}
& \shortstack{OEM-SAR\\mIoU $\uparrow$}
& \shortstack{DIOR-R\\mAP50 $\uparrow$} \\
\midrule
\multicolumn{8}{l}{\textit{Baselines evaluated on multiple datasets}} \\
\methodsrc{zhang2026unichange}{UniChange}{CVPR'26}   & \cellcolor{gray!16}90.41 & \cellcolor{gray!16}\textbf{78.87} & & & & & \\
\methodsrc{guo2024skysense}{SkySense}{CVPR'24}    & & & \cellcolor{gray!16}59.30 & & & & \cellcolor{gray!16}74.27 \\
\methodsrc{wang2024mtp}{MTP}{JSTARS'24}         & & & & \cellcolor{gray!16}54.17 & & & \cellcolor{gray!16}74.54 \\
\methodsrc{wang2025selectivemae}{SelectiveMAE}{ICCV'25} & & & & \cellcolor{gray!16}53.92 & & & \cellcolor{gray!16}71.75 \\
\methodsrc{tan2025fedsense}{FedSense}{ICCV'25} & & & & \cellcolor{gray!16}55.96 & & & \cellcolor{gray!16}68.82 \\
\methodsrc{lv2026s5}{S5}{AAAI'26}       & & & & \cellcolor{gray!16}55.65 & \cellcolor{gray!16}70.02 & & \cellcolor{gray!16}75.30 \\
\methodsrc{wang2024metasegnet}{MetaSegNet}{TGRS'24}  & & & & \cellcolor{gray!16}52.00 & \cellcolor{gray!16}70.40 & & \\
\midrule
\multicolumn{8}{l}{\textit{Source-specific baselines}} \\
\methodsrc{li2025uabcd}{UA-BCD}{ISPRS JPRS'25}             & \cellcolor{gray!16}86.87 & & & & & & \\
\methodsrc{zhou2026tricdnet}{TriCDNet}{ISPRS Arch.'26}       & \cellcolor{gray!16}86.91 & & & & & & \\
\methodsrc{yang2026changerwkv}{ChangeRWKV-B}{AAAI'26}        & & \cellcolor{gray!16}75.46 & & & & & \\
\methodsrc{dong2024changeclip}{ChangeCLIP}{ISPRS JPRS'24}     & & \cellcolor{gray!16}75.63 & & & & & \\
\methodsrc{cheng2022mask2former}{Mask2Former}{CVPR'22}        & & & \cellcolor{gray!16}\textbf{61.06} & & & & \\
\methodsrc{zou2025d2ls}{D2LS}{ICCV'25}                        & & & & \cellcolor{gray!16}55.30 & & & \\
\methodsrc{xia2023openearthmap}{FT-U-NetFormer}{WACV'23}      & & & & & \cellcolor{gray!16}69.13 & & \\
\methodsrc{xia2025openearthmapsar}{VMamba}{GRSM'25}           & & & & & & \cellcolor{gray!16}34.74 & \\
\methodsrc{li2025segearthov}{SegEarth-OV}{CVPR'25}            & & & & & & \cellcolor{gray!16}35.10 & \\
\methodsrc{xia2025openearthmapsar}{U-Net-EffNet-B4}{GRSM'25}  & & & & & & \cellcolor{gray!16}35.13 & \\
\methodsrc{xia2025openearthmapsar}{SegFormer}{GRSM'25}         & & & & & & \cellcolor{gray!16}\textbf{35.77} & \\
\methodsrc{huang2024mutdet}{MutDet}{ECCV'24}                  & & & & & & & \cellcolor{gray!16}70.70 \\
\methodsrc{lee2025hausdorff}{RHINO-DETR}{WACV'25}             & & & & & & & \cellcolor{gray!16}72.67 \\
\methodsrc{bi2025ringmoe}{RingMoE}{TPAMI'26}                  & & & & & & & \cellcolor{gray!16}\textbf{76.04} \\
\midrule
\textbf{\method{}}
& \textbf{90.88} & 76.44 & 56.57 & \textbf{56.30} & \textbf{71.28} & 30.12 & 64.75 \\
\bottomrule
\end{tabular}
\end{table}

\begin{table}[H]
\caption{
Benchmark comparison for building extraction, height estimation, and referring segmentation. 
}
\label{tab:main_b}
\centering
\tablebodyfont
\setlength{\tabcolsep}{4.0pt}
\renewcommand{\arraystretch}{1.00}
\begin{tabular}{lcccccccc}
\toprule
& \multicolumn{2}{>{\columncolor{taskBuilding}}c}{Building extraction}
& \multicolumn{2}{>{\columncolor{taskHeight}}c}{Height estimation}
& \multicolumn{4}{>{\columncolor{taskReferring}}c}{Referring segmentation} \\
\cmidrule(lr){2-3}\cmidrule(lr){4-5}\cmidrule(lr){6-9}
Method
& \shortstack{WHU\\IoU $\uparrow$}
& \shortstack{SpaceNet 6\\SAR IoU $\uparrow$}
& \shortstack{Potsdam\\RMSE $\downarrow$}
& \shortstack{DFC23\\$\delta_1 \uparrow$}
& \shortstack{RRSIS-D\\cIoU $\uparrow$}
& \shortstack{RRSIS-D\\gIoU $\uparrow$}
& \shortstack{LaSeRS\\cIoU $\uparrow$}
& \shortstack{LaSeRS\\gIoU $\uparrow$} \\
\midrule
\methodsrc{zhao2025ccs}{CCS (SAM2)}{CVPR'25}                & \cellcolor{gray!16}88.58 & & & & & & & \\
\methodsrc{shi2021dmbc}{DMBC-Net}{Remote Sens.'21}                   & \cellcolor{gray!16}91.66 & & & & & & & \\
\methodsrc{yao2026uaglnet}{UAGLNet}{TGRS'26}                            & \cellcolor{gray!16}92.07 & & & & & & & \\
\methodsrc{he2024uanet}{UANet}{TGRS'24}                              & \cellcolor{gray!16}\textbf{92.15} & & & & & & & \\
\methodsrc{wang2024bidnet}{BIDNet}{TGRS'24}                          & & \cellcolor{gray!16}55.20 & & & & & & \\
\methodsrc{wei2026cmhcl}{CMH-CL}{JSTARS'26}                          & & \cellcolor{gray!16}55.25 & & & & & & \\
\methodsrc{pei2026aemkd}{AEM-KD}{JSTARS'26}                          & & \cellcolor{gray!16}55.43 & & & & & & \\
\methodsrc{chen2024heightformer}{HeightFormer}{Remote Sens.'24}       & & & \cellcolor{gray!16}2.193 & & & & & \\
\methodsrc{du2024lumnet}{LUMNet}{GRSL'24}                            & & & \cellcolor{gray!16}2.158 & & & & & \\
\methodsrc{lei2025raephrnet}{RAEPHR-Net}{JSTARS'25}                  & & & \cellcolor{gray!16}\textbf{2.018} & & & & & \\
\methodsrc{xiong2024dofa}{DoFA}{arXiv'24}                             & & & & \cellcolor{gray!16}30.99 & & & & \\
\methodsrc{mao2023light}{LIGHT}{IGARSS'23}                            & & & & \cellcolor{gray!16}51.10 & & & & \\
\methodsrc{yang2026mars}{MaRS}{AAAI'26}                  & & & & \cellcolor{gray!16}54.04 & & & & \\
\methodsrc{liu2024rmsin}{RMSIN}{CVPR'24}                             & & & & & \cellcolor{gray!16}77.79 & \cellcolor{gray!16}64.20 & & \\
\methodsrc{rong2026rs2sam2}{RS2-SAM2}{AAAI'26}                      & & & & & \cellcolor{gray!16}78.99 & \cellcolor{gray!16}66.72 & & \\
\methodsrc{jang2025mmr}{M$^2$A-7B}{ICLR'25}                         & & & & & & & \cellcolor{gray!16}29.30 & \cellcolor{gray!16}30.50 \\
\methodsrc{shabbir2025geopixel}{GeoPixel-8B}{ICML'25}               & & & & & & & \cellcolor{gray!16}55.20 & \cellcolor{gray!16}50.40 \\
\methodsrc{xin2026segearthr2}{SegEarth-R2-3B}{CVPR'26}                 & & & & & & \cellcolor{gray!16}67.90 & \cellcolor{gray!16}67.90 & \cellcolor{gray!16}57.20 \\
\midrule
\textbf{\method{}}
& 89.14 & \textbf{64.72} & 2.121 & \textbf{73.31} & \textbf{80.11} & \textbf{70.33} & \textbf{82.13} & \textbf{77.53} \\
\bottomrule
\end{tabular}
\vspace{1pt}

\end{table}

\FloatBarrier
\Needspace{8\baselineskip}
\subsection{Comparison with General-Purpose Generative Prediction}
\label{app:instructcv_comparison}

\rev{General-purpose generative frameworks offer a promising
approach to unified visual prediction, but extending them
to remote sensing requires accommodating heterogeneous
sensing modalities, input configurations, and output structures.
We select InstructCV~\citep{gan2024instructcv}, an
instruction-conditioned diffusion framework, as a
representative baseline to investigate this challenge.}

\rev{\paragraph{Implementation details.}
We adapt InstructCV to the same six tasks and thirteen
datasets while retaining its original Stable Diffusion
backbone, image--text conditioning, and diffusion objective,
without introducing TMAS or replacing its backbone with FLUX.
The adaptation extends its prediction interface to support
optical and SAR imagery, bi-temporal observations, continuous
height maps, and rotated instances through corresponding
visual target representations and recovery procedures.}

\rev{A single InstructCV model is jointly trained on all
thirteen datasets using image--instruction--target triplets,
following the same data splits and sampling proportions
as \method{}.
The model is initialized from its official pretrained
checkpoint and trained with optimization settings appropriate
to its diffusion architecture.
Training budgets are matched by effective batch size and
per-source sample exposure, with GPU-hours reported separately.
Inference follows the official settings: $512\times512$
resolution, 100 sampling steps, text guidance of 7.5,
and image guidance of 1.5.
Both methods are evaluated on identical test splits using
the same valid regions and dataset-native metrics.
Checkpoint selection and evaluation configurations are
determined without using test results.}

\begin{table}[H]
\caption{
\rev{Comparison with InstructCV across six tasks and thirteen datasets.
All metrics are percentages except Potsdam RMSE (m).
Referring-segmentation results are reported as cIoU/gIoU.}
}
\label{tab:instructcv_comparison}
\centering
\tablebodyfont
\setlength{\tabcolsep}{5pt}
\renewcommand{\arraystretch}{1.12}
\begin{tabular}{@{}lllcc@{}}
\toprule
Task & Dataset & Metric
& \methodsrc{gan2024instructcv}{InstructCV}{ICLR'24}
& \method{} \\
\midrule
\multirow{2}{*}{Building extraction}
& WHU Building & IoU $\uparrow$
& 81.06 & \textbf{89.14} \\
& SpaceNet 6 (SAR) & IoU $\uparrow$
& 48.39 & \textbf{64.72} \\
\midrule
\multirow{3}{*}{Change detection}
& WHU-CD & IoU $\uparrow$
& 83.27 & \textbf{90.88} \\
& LEVIR-CD+ & IoU $\uparrow$
& 61.07 & \textbf{76.44} \\
& UrbanSARFloods & mIoU $\uparrow$
& 42.16 & \textbf{56.57} \\
\midrule
\multirow{3}{*}{Semantic segmentation}
& LoveDA & mIoU $\uparrow$
& 42.04 & \textbf{56.30} \\
& OpenEarthMap-RGB & mIoU $\uparrow$
& 58.70 & \textbf{71.28} \\
& OpenEarthMap-SAR & mIoU $\uparrow$
& 14.72 & \textbf{30.12} \\
\midrule
\multirow{2}{*}{Height estimation}
& Potsdam & RMSE $\downarrow$
& 2.875 & \textbf{2.121} \\
& DFC23 Track 2 & $\delta_1$ $\uparrow$
& 36.82 & \textbf{73.31} \\
\midrule
\multirow{2}{*}{Referring segmentation}
& RRSIS-D & cIoU/gIoU $\uparrow$
& 73.17 / 64.27
& \textbf{80.11} / \textbf{70.33} \\
& LaSeRS & cIoU/gIoU $\uparrow$
& 69.35 / 63.04
& \textbf{82.13} / \textbf{77.53} \\
\midrule
Rotated detection & DIOR-R & mAP50 $\uparrow$
& 40.28 & \textbf{64.75} \\
\bottomrule
\end{tabular}
\end{table}

\rev{\paragraph{Results.}
As shown in Table~\ref{tab:instructcv_comparison},
\method{} outperforms the adapted InstructCV across all
thirteen benchmarks, spanning diverse sensing modalities
and output structures.
Notable gains include 16.33 IoU points on SpaceNet 6 SAR,
15.40 mIoU points on OpenEarthMap-SAR, and 24.47 mAP50
points on DIOR-R.
For height estimation, \method{} improves DFC23
$\delta_1$ from 36.82 to 73.31 and reduces Potsdam
RMSE from 2.875 to 2.121 m.
These results highlight the challenges of extending
general-purpose generative prediction to heterogeneous
remote-sensing tasks and demonstrate the effectiveness
of a unified prediction framework designed for
remote-sensing perception.}

\clearpage
\section{Additional Qualitative Comparisons}
\label{app:additional_qualitative}

\subsection{Comparisons across the Six Task Families}

\begin{figure}[H]
\centering
\includegraphics[width=\linewidth,height=0.55\textheight,keepaspectratio]{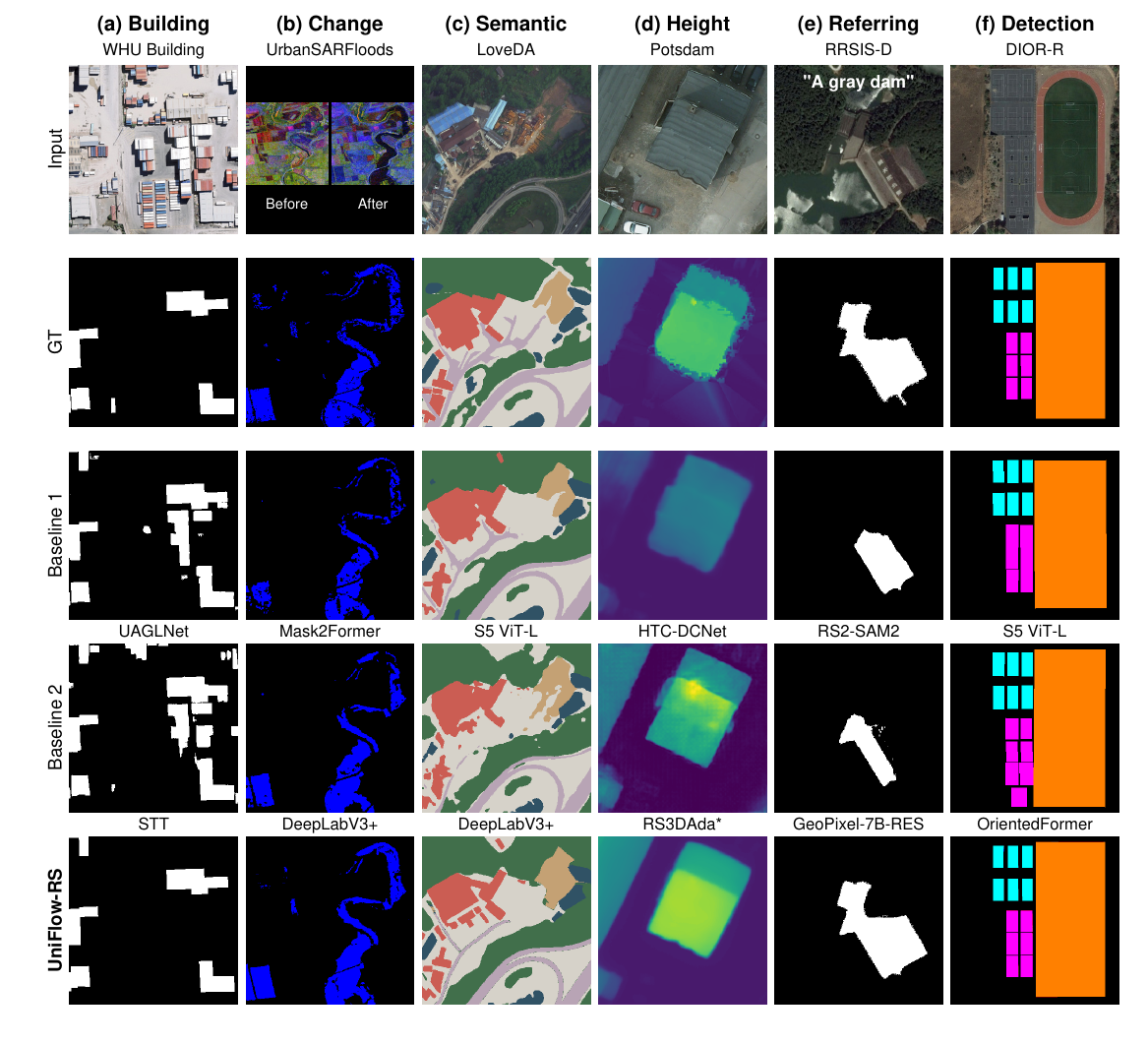}
\caption{\textbf{Additional qualitative comparisons across six tasks.}
Rows show the input, ground truth, two baseline predictions, and \method{}.
Baseline methods are labeled below each prediction and cited in the text.}
\label{fig:additional_qualitative}
\end{figure}

Figure~\ref{fig:additional_qualitative} provides additional examples across all
six task families, complementing the numerical results in
Appendix~\ref{app:full_benchmarks}. Each column shows the same input and target
with two representative baselines and \method{}.

\paragraph{Compared methods.}
We compare with UAGLNet~\citep{yao2026uaglnet} and STT~\citep{chen2021stt}
for building extraction; Mask2Former~\citep{cheng2022mask2former} and
DeepLabV3+~\citep{chen2018deeplabv3plus} for flood change mapping;
S5 ViT-L~\citep{lv2026s5} and DeepLabV3+ for LoveDA;
HTC-DCNet~\citep{htcdc} and RS3DAda~\citep{song2024synrs3d} for height estimation;
RS2-SAM2~\citep{rong2026rs2sam2} and GeoPixel-7B-RES~\citep{shabbir2025geopixel}
for referring segmentation; and S5 ViT-L and OrientedFormer-R50~\citep{zhao2024orientedformer}
for rotated detection.

\paragraph{Visual observations.}
\method{} shows fewer spurious responses in building extraction, follows the
curved flooded region in SAR change mapping, and recovers more of the queried
gray dam in referring segmentation. It also preserves clearer roof-height
variation and separation between adjacent sports facilities. Semantic differences
mainly occur around building and land-cover boundaries. Overall, the shared
predictor preserves task-specific spatial structures across heterogeneous tasks.

\clearpage
\subsection{Referring Segmentation on Unseen Datasets}
\label{app:qualitative_referring_transfer}

\begin{figure}[H]
\centering
\includegraphics[width=\linewidth]
{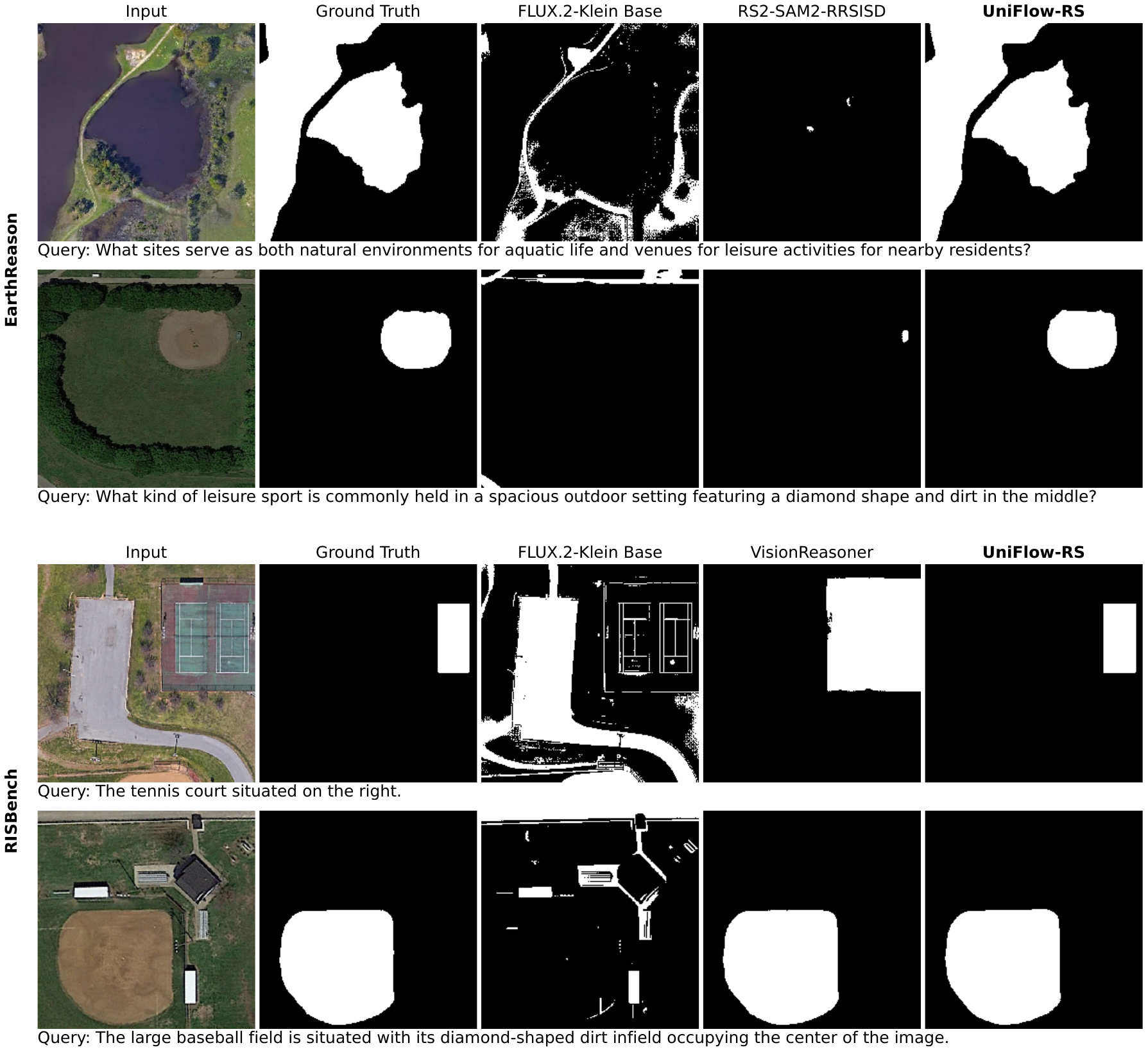}
\caption{
\rev{Qualitative zero-shot referring-segmentation results
on EarthReason and RISBench.
Each row shows the same image, query, ground truth,
and model predictions.
Comparisons include FLUX.2-Klein Base,
RS2-SAM2-RRSISD~\citep{rong2026rs2sam2}
on EarthReason, and
VisionReasoner~\citep{liu2026visionreasoner}
on RISBench.
White denotes foreground.
VisionReasoner predictions are reproduced using
its released checkpoint and official pipeline
on the selected examples.}
}
\label{fig:qualitative_referring_transfer}
\end{figure}

\rev{Figure~\ref{fig:qualitative_referring_transfer}
presents qualitative zero-shot transfer results on
EarthReason and RISBench, complementing the quantitative
comparisons in Tables~\ref{tab:generalization}
and~\ref{tab:generalization_full}.}

\rev{On EarthReason, \method{} successfully identifies
water bodies and a baseball field from indirect,
reasoning-oriented descriptions.
On RISBench, it accurately distinguishes the
right-hand tennis court from adjacent objects,
while the baseball example shows close agreement
with VisionReasoner.
These examples illustrate the ability of the shared
generative predictor to adapt to unseen visual
distributions and diverse language expressions
without target-specific adaptation.}

\clearpage
\subsection{Occlusion and Spectral-Composition Transfer}
\label{app:qualitative_occlusion_irrg}

\begin{figure}[H]
\centering
\includegraphics[width=\linewidth,height=0.68\textheight,keepaspectratio]
{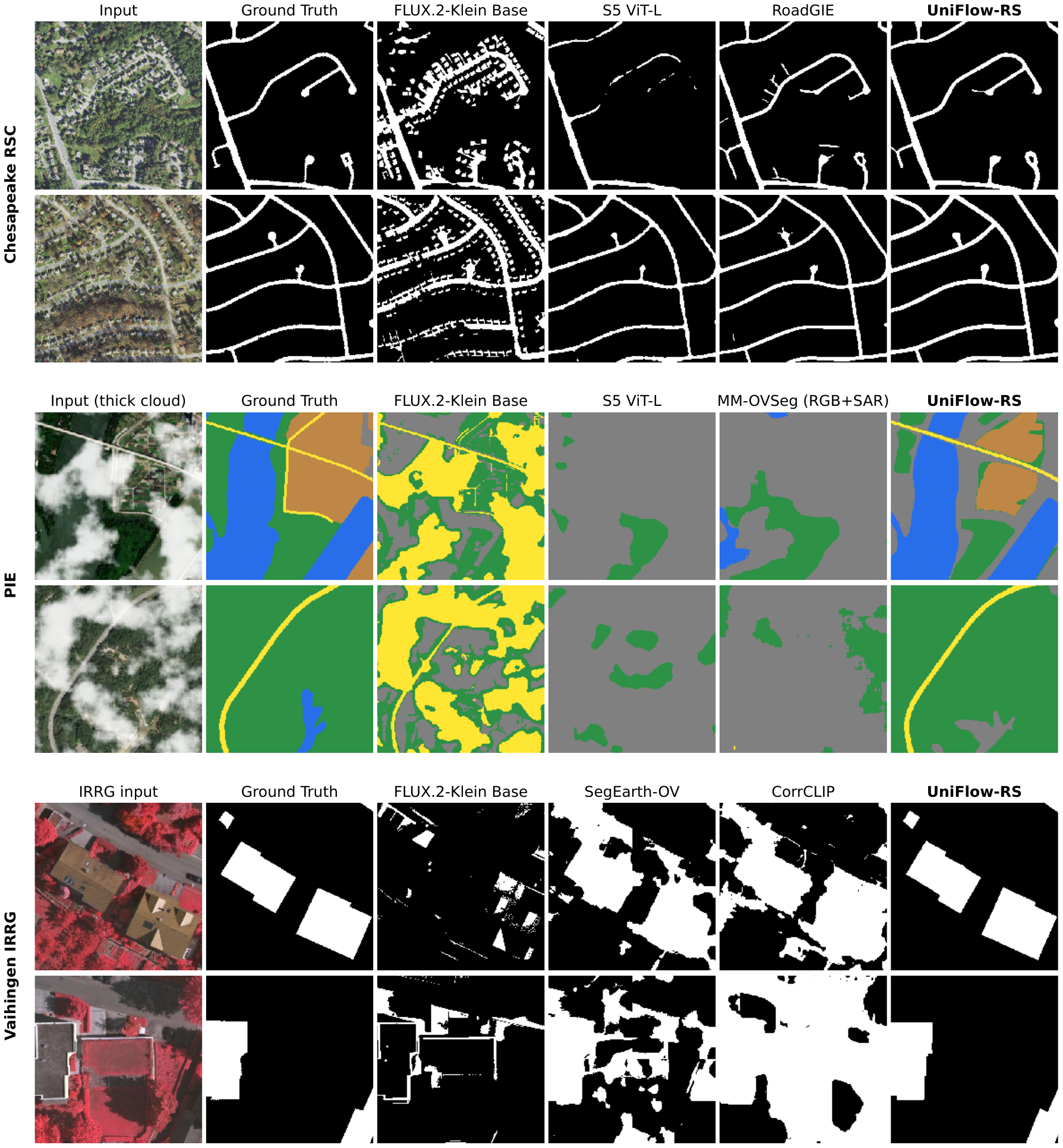}
\caption{
\rev{Qualitative zero-shot transfer under occlusion
and spectral shifts.
Two examples are shown for each dataset.
Comparisons include S5~\citep{lv2026s5},
the RoadGIE automatic baseline~\citep{peng2026roadgie},
MM-OVSeg~\citep{wei2026mmovseg},
SegEarth-OV~\citep{li2025segearthov},
and CorrCLIP~\citep{zhang2025corrclip}.
PIE uses synthetic thick-cloud conditions,
with MM-OVSeg additionally receiving uncorrupted SAR.
Semantic colors indicate road (yellow), water (blue),
forest (green), farmland (brown), and Other (gray);
binary foreground is shown in white.}
}
\label{fig:qualitative_occlusion_irrg}
\end{figure}

\rev{Figure~\ref{fig:qualitative_occlusion_irrg}
visualizes zero-shot transfer under canopy occlusion,
cloud corruption, and spectral shifts, complementing
Tables~\ref{tab:rsc_occlusion}
and~\ref{tab:vaihingen_irrg}.}

\rev{On Chesapeake RSC, \method{} recovers more continuous
road structures beneath tree canopies.
On PIE, it preserves road and land-cover regions
under synthetic cloud occlusion.
On Vaihingen IRRG imagery, the model produces
more complete building footprints with fewer
surrounding false positives.
Together, these examples illustrate the ability
of the shared generative predictor to preserve
spatial structures across unseen task settings
and challenging observation conditions.}

\end{document}